\documentclass[acmtog]{acmart}
\AtBeginDocument{%
  }

\setcopyright{none}
\renewcommand\footnotetextcopyrightpermission[1]{}
\usepackage{multirow} % 引入 multirow 宏包
\usepackage{xcolor} % 颜色
\usepackage{subcaption}
\usepackage{caption}
\usepackage{hyperref}

\definecolor{mcolor}{HTML}{B556D8}   % 粉色 (M)
\definecolor{acolor}{HTML}{877DCC}   % 紫色 (A)
\definecolor{ccolor}{HTML}{79ABD3}   % 蓝色 (C)
\definecolor{ecolor}{HTML}{82D386}   % 绿色 (E)

\begin{document}

\title{\textbf{\textcolor{mcolor}{M}\textcolor{acolor}{A}\textcolor{ccolor}{C}\textcolor{ecolor}{E}\textcolor{black}{-Dance:}
    \textcolor{mcolor}{Motion}-\textcolor{acolor}{Appearance} \textcolor{ccolor}{Cascaded} \textcolor{ecolor}{Experts} for Music-Driven Dance Video Generation
  }
}

\author{Kaixing Yang}
\authornote{Work done during an internship at AMap, Alibaba Group.}
\email{yangkaixing@ruc.edu.cn}
\affiliation{
  \institution{Renmin University of China}
  \city{Beijing}
  \country{China}
}

\author{Jiashu Zhu}
\authornote{Project leader.}
\email{zhujiashu.zjs@alibaba-inc.com}
\affiliation{
  \institution{AMAP, Alibaba Group}
  \city{Beijing}
  \country{China}
}

\author{Xulong Tang}
\email{xulong.tang@maloutech.com}
\affiliation{
  \institution{Malou Tech Inc}
  \city{Richardson}
  \state{Texas}
  \country{USA}
}

\author{Ziqiao Peng}
\email{pengziqiao@ruc.edu.cn}
\affiliation{
  \institution{Renmin University of China}
  \city{Beijing}
  \country{China}
}

\author{Xiangyue Zhang}
\email{xiangyuezhang@whu.edu.cn}
\affiliation{
  \institution{Wuhan University}
  \city{Wuhan}
  \country{China}
}

\author{Puwei Wang}
\authornote{Corresponding authors.}
\email{wangpuwei@ruc.edu.cn}
\affiliation{
  \institution{Renmin University of China}
  % \city{Beijing}
  \country{China}
}

\author{Jiahong Wu}
\authornotemark[3]
\email{hongxi.wjh@alibaba-inc.com}
\affiliation{
  \institution{AMAP, Alibaba Group}
  \city{Beijing}
  \country{China}
}

\author{Xiangxiang Chu}
\email{cxxgtxy@gmail.com}
\affiliation{
  \institution{AMAP, Alibaba Group}
  \city{Beijing}
  \country{China}
}

\author{Hongyan Liu}
\authornotemark[3]
\email{liuhy@sem.tsinghua.edu.cn}
\affiliation{
  \institution{Tsinghua University}
  \city{Beijing}
  \country{China}
}

\author{Jun He}
% \authornotemark[2]
\email{hejun@ruc.edu.cn}
\affiliation{
  \institution{Renmin University of China}
  \city{Beijing}
  \country{China}
}

\begin{abstract}
With the rise of online dance‑video platforms and rapid advances in AIGC, music‑driven dance generation task has emerged as a compelling research direction. 
Despite substantial progress in related domains such as music-driven 3D dance generation, pose-driven image animation, and audio-driven talking-head synthesis, these approaches are not readily transferable to this task due to fundamental mismatches in generation targets and constraints. Moreover, research on music-driven dance video generation remains limited and fails to capture the inherently 3D nature of dance, resulting in compromised motion quality and visual appearance.
Accordingly, we present \textbf{MACE-Dance}, a music-driven dance video generation framework with cascaded Mixture-of-Experts (MoE). The Motion Expert performs music-to-3D motion enforcing kinematic plausibility and artistic expressiveness, while the Appearance Expert carries out motion-and-reference conditioned video synthesis, preserving visual identity with spatiotemporal coherence. Specifically, the \textbf{Motion Expert} adopts Diffusion Model with BiMamba-Transformer hybrid architecture and Guidance-Free Training (GFT) strategy, achieving state-of-the-art (SOTA) performance in 3D dance generation task; the \textbf{Appearance Expert} adopts a decoupled Kinematic–Aesthetic fine-tuning strategy, achieving state-of-the-art (SOTA) performance in the pose-driven image animation task. To better benchmark this task, we curate a large-scale dataset, and design a motion–appearance evaluation protocol. Based on them, \textbf{MACE-Dance} also achieves the state-of-the-art (SOTA) performance. Code is available at \url{https://github.com/AMAP-ML/MACE-Dance}.

\end{abstract}



% \keywords{AI for Art, Multimodal Learning, Digital Human Generation, Video Generation}

\maketitle

\begin{figure*}
  \centering
  \includegraphics[width=0.9\textwidth]{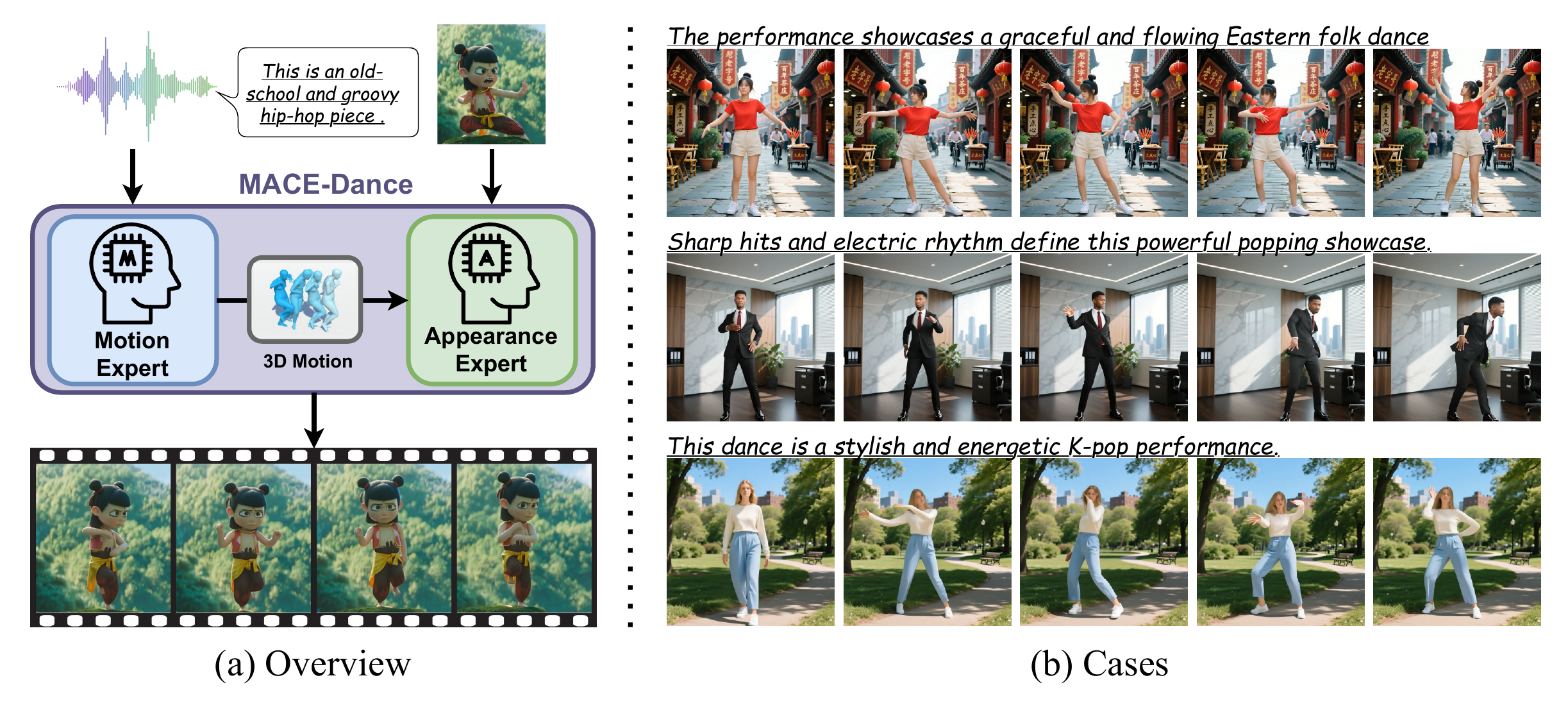}
  % \vspace{-0.2in}
  \caption{Leveraging the synergistic collaboration among the cascaded experts, \textbf{MACE-Dance} can generate diverse dance videos that not only exhibit kinematically plausible and artistically expressive motion, but also maintain spatiotemporal coherent appearance.}
  % \vspace{-0.05in}
  \label{fig: teaser}
\end{figure*}

\section{Introduction}
Dance is a vital part of human culture. Moving to the beat and melody, dancers both convey emotion and narrative intent and showcase the power and beauty of human movement~\cite{tseng2023edge,butterworth2004teaching}. In the era of the internet, dance videos have become highly prominent on platforms such as YouTube and TikTok. In parallel, rapid advances~\cite{yang2025matchdance,zhuo2023video,chen2025taming,chen2025s,lei2025there} in AI-generated content (AIGC) has created the technical preconditions for automating dance video creation, making it a timely and impactful research direction. Nevertheless, this task faces two key challenges: (1) generating dance \textbf{motions} that are kinematically plausible while artistically expressive; and (2) achieving high-fidelity visual \textbf{appearance} with strong spatiotemporal consistency.

Recent progress in dance generation has focused primarily on 3D dance~\cite{tseng2023edge,li2024lodge,li2023finedance}, with numerous strong methods emerging across model families-autoregressive~\cite{siyao2022bailando,yang2025megadance,yang2025matchdance}, GAN-based~\cite{yang2024cohedancers,sun2019deep,huang2021choreography}, and diffusion-based~\cite{tseng2023edge,li2024lodge,li2024lodge++}. Although 2D dance videos can be rendered from 3D motion, such renderings typically lack realistic human–scene interactions and detailed appearance cues, resulting in visually suboptimal outputs~\cite{yang2024beatdance}.
In contrast, human-centric image animation leverages a reference image along with various driving signals to generate videos. 
Traditionally, pose-driven image animation has achieved notable advances~\cite{tan2024animate,hu2024animate,cheng2025wan}. However, its utility for dance video generation is limited, as pose design—widely regarded as the most challenging and time-consuming step—still remains manual~\cite{butterworth2004teaching}.
Similarly, audio-driven talking head generation has also achieved significant breakthroughs~\cite{peng2024synctalk,peng2025synctalk++,peng2025omnisync}. However, its direct transfer to dance video generation remains challenging, as it primarily focuses on relatively simple upper-body gesture rather than the complex full-body motion required in dance~\cite{peng2023emotalk}. 
Research on music-driven dance video generation remains limited~\cite{chen2025x,wang2025dance,tang2025spatial} and fails to capture the inherently 3D nature of dance, resulting in compromised motion quality and visual appearance.

Accordingly, we present \textbf{MACE-Dance}, a music-driven dance video generation framework with cascaded mixture-of-experts (MoE), as shown in Fig. ~\ref{fig: teaser}. The Motion Expert performs music-to-3D motion enforcing kinematic plausibility and artistic expressiveness, while the Appearance Expert carries out motion-and-reference conditioned video synthesis, preserving visual identity with spatiotemporal coherence. Notably, \textbf{MACE-Dance} adopts 3D SMPL~\cite{loper2023smpl} parameters rather than 2D keypoints as the intermediate representation, as 3D provides view-invariant and physically consistent supervision, while 2D projections introduce irreversible information loss and viewpoint ambiguity. \textbf{(1) Motion Expert.} Motion Expert adopts Diffusion Model with BiMamba-Transfomer hybrid architecture. The bidirectional Mamba~\cite{gu2023mamba} captures intra-modal local dependencies in music or dance, while the Transformer~\cite{vaswani2017attention} models cross-modal global context. Owing to this architecture, the Motion Expert generates entire sequence in non-autoregressive manner during inference, not only improving generation efficiency, but also avoiding the exposure bias problem in autoregressive~\cite{yang2025megadance} and inpainting-based~\cite{tseng2023edge} methods. To enhance generation stability and accelerate inference, we employ guidance-free training (GFT~\cite{chen2025visual}) instead of conventional classifier-free guidance (CFG~\cite{ho2022classifier}), enhancing the physical plausibility and artistic expressiveness for the generated dance. 
\textbf{(2) Appearance Expert.} Wan-Animate~\cite{cheng2025wan} has recently garnered substantial attention in both industry and academia. However, directly applying it to dance video generation yields limited effectiveness, as dance videos exhibit significantly more complex patterns than general videos. Thus, the Appearance Expert adopts a decoupled Kinematic–Aesthetic two-stage fine-tuning strategy to achieve high-fidelity appearance synthesis. In Kinematic stage, it fine-tunes the Body Adapter to strengthen kinematic conditioning and motion adherence. In Aesthetic stage, it attaches a LoRA~\cite{hu2022lora} branch to each DiT block and fine-tunes for aesthetic refinement, enhancing texture fidelity and stylistic consistency.
 
To better benchmark music-driven dance video generation task, we curate a large-scale dataset and design a motion–appearance evaluation protocol. Firstly, we curate a large-scale dance video dataset, named MA-Data, comprising ~70k clips of 5–10 seconds each (totaling 116 hours), spanning over 20 dance genres. The dataset consists of two complementary sources: (1) 3D-rendered data (motion-centric): Derived from FineDance~\cite{li2023finedance}—the largest 3D dance dataset recorded by professional dancers—we render front-view videos and extract random 5–10 s segments via a sliding window, yielding ~20k clips (28 h). This subset emphasizes professional dance motion rather than visual appearance. (2) In-the-wild internet data (appearance-centric): Collected from high-engagement videos on platforms such as TikTok and YouTube, using the same sliding-window strategy to obtain ~50k 5-10 s clips (88 h). This subset emphasizes visual appearance, while motions are relatively unprofessional. Secondly, we introduce a motion–appearance evaluation protocol. For the motion dimension, we assess the fidelity, diversity, and synchronization~\cite{li2021ai,li2023finedance} from Human-Kinematics perspective based on the 2D keypoints extracted by ViTPose~\cite{xu2022vitpose}. For the appearance dimension, we adopt VBench~\cite{huang2024vbench}—a widely used benchmark in video generation—and select a set of dance-specific metrics.

In conclusion, our contributions are as follows:
(1) To better benchmark the music-driven dance video generation task, we curate a large-scale dataset named \textbf{MA-Data}, along with a motion–appearance evaluation protocol.
(2) Based on them, we introduce \textbf{MACE-Dance}, a music-driven dance video generation framework with cascaded experts, achieving SOTA performance.
(3) The \textbf{Motion Expert} adopts Diffusion Model with BiMamba-Transformer hybrid architecture and Guidance-Free Training strategy, achieving SOTA performance on the FineDance dataset in music-driven 3D dance generation task.
(4) \textbf{Appearance Expert} adopts a decoupled Kinematic-Aesthetic fine-tuning strategy, achieving SOTA performance on the MA-Data dataset in the pose-driven image animation task.

% \begin{itemize}
%     \item To better benchmark the music-driven dance video generation task, we curate a large-scale dataset named \textbf{MA-Data}, along with a motion–appearance evaluation protocol.
    
%     \item Based on them, we introduce \textbf{MACE-Dance}, a music-driven dance video generation framework with cascaded experts, achieving state-of-the-art (SOTA) performance.
    
%     \item The \textbf{Motion Expert} adopts Diffusion Model with BiMamba-Transformer hybrid architecture and Guidance-Free Training strategy, achieving SOTA performance on the FineDance dataset in music-driven 3D dance generation task.
    
%     \item The \textbf{Appearance Expert} adopts a decoupled Kinematic-Aesthetic fine-tuning strategy, achieving SOTA performance on the MA-Data dataset in the pose-driven image animation task.
% \end{itemize}

\section{Related Work}
\subsection{Music-Driven 3D Dance Generation}
Music and dance are deeply intertwined, and recent progress in music-to-dance generation has largely centered on 3D motion. Broadly, existing methods fall into three families: GAN-based, autoregressive, and diffusion-based models.
\textbf{1) GAN-based models.} Generators synthesize motion from music while discriminators provide adversarial feedback. Examples include CoheDancers~\cite{yang2024cohedancers} and DeepDance~\cite{sun2019deep}. \textbf{2) Autoregressive models.} These methods typically adopt a two-stage pipeline: curating choreographic units by VQ-VAE~\cite{oord2017neural} or FSQ~\cite{mentzer2023finite}, followed by autoregressive modeling of music-conditioned distributions over these units~\cite{yang2024codancers,yang2026tokendance,li2024exploring}. Works such as Bailando~\cite{siyao2022bailando}, Bailando++~\cite{siyao2023bailando++}, and MEGADance~\cite{yang2025megadance} fall into this paradigm. \textbf{3) Diffusion-based models.} These methods corrupt motion with noise and train denoising networks to iteratively recover sequences conditioned on music~\cite{yang2025flowerdance}, enabling diverse and temporally coherent dances. Representative works include EDGE~\cite{tseng2023edge}, FineNet~\cite{li2023finedance}, Lodge~\cite{li2024lodge}, Lodge++~\cite{li2024lodge++}, and GCDance~\cite{liu2025gcdance}. Despite substantial progress, 3D dance generation only focuses on motion generation and underemphasizes visual appearance—an essential aspect of dance as an art form. Although 2D dance videos can be rendered from 3D motion, the outputs typically lack realistic human–scene interactions and high-fidelity human textures.

\subsection{Human-Centric Image Animation}
In contrast, human-centric image animation leverages a reference image along with various driving signals to generate videos that exhibit high-quality visual appearance, making it a promising direction for dance video generation.
Firstly, pose-driven image animation utilizes 2D keypoints to generate motion videos, achieving notable advances~\cite{tan2024animate,hu2024animate,cheng2025wan}, including Animate-X~\cite{tan2024animate}, Animate Anyone~\cite{hu2024animate} and Wan-Animate~\cite{cheng2025wan}. However, its utility for dance video generation is limited, as pose design—widely regarded as the most challenging and time-consuming step— still remains manual~\cite{butterworth2004teaching}.
Secondly, speech-driven image animation employs audio features to generate talking head videos, also achieving significant breakthroughs~\cite{peng2024synctalk,peng2025synctalk++,peng2025omnisync}, such as SyncTalk~\cite{peng2024synctalk}, OmniSync~\cite{peng2025omnisync} and Hallo2~\cite{cui2024hallo2}. However, its direct transfer to dance video generation remains challenging, as these methods primarily focus on relatively simple upper-body gestures rather than the complex full-body motion required in dance~\cite{peng2023emotalk,zhang2025semtalk,zhang2025mitigating,zhang2025echomask}. 
Finally, research on music-driven dance video generation remains limited. DabFusion~\cite{wang2025dance} introduces an end-to-end Diffusion-based method, but the generated videos exhibit blurry foreground subjects and backgrounds, thereby degrading visual fidelity. X-Dancer ~\cite{chen2025x}, STG-Mamba~\cite{tang2025spatial} and ChoreoMuse~\cite{wang2025choreomuse} predict 2D keypoints from music and then drives image animation with these keypoints. However, they remain limited in handling limb occlusions and complex full-body locomotion in dance videos.
In conclusion, existing works for dance video generation still fails to capture the inherently 3D nature of dance, resulting in compromised motion quality and visual appearance. Thus, we propose \textbf{MACE-Dance}, a cascaded expert framework that synergistically integrates motion and appearance generation, producing kinematically plausible and artistically expressive motion while maintaining spatiotemporally coherent visual appearance.

\section{Methodology}
\subsection{Overview}
\begin{figure*}[t]
    \centering
    \includegraphics[width=0.9\linewidth]{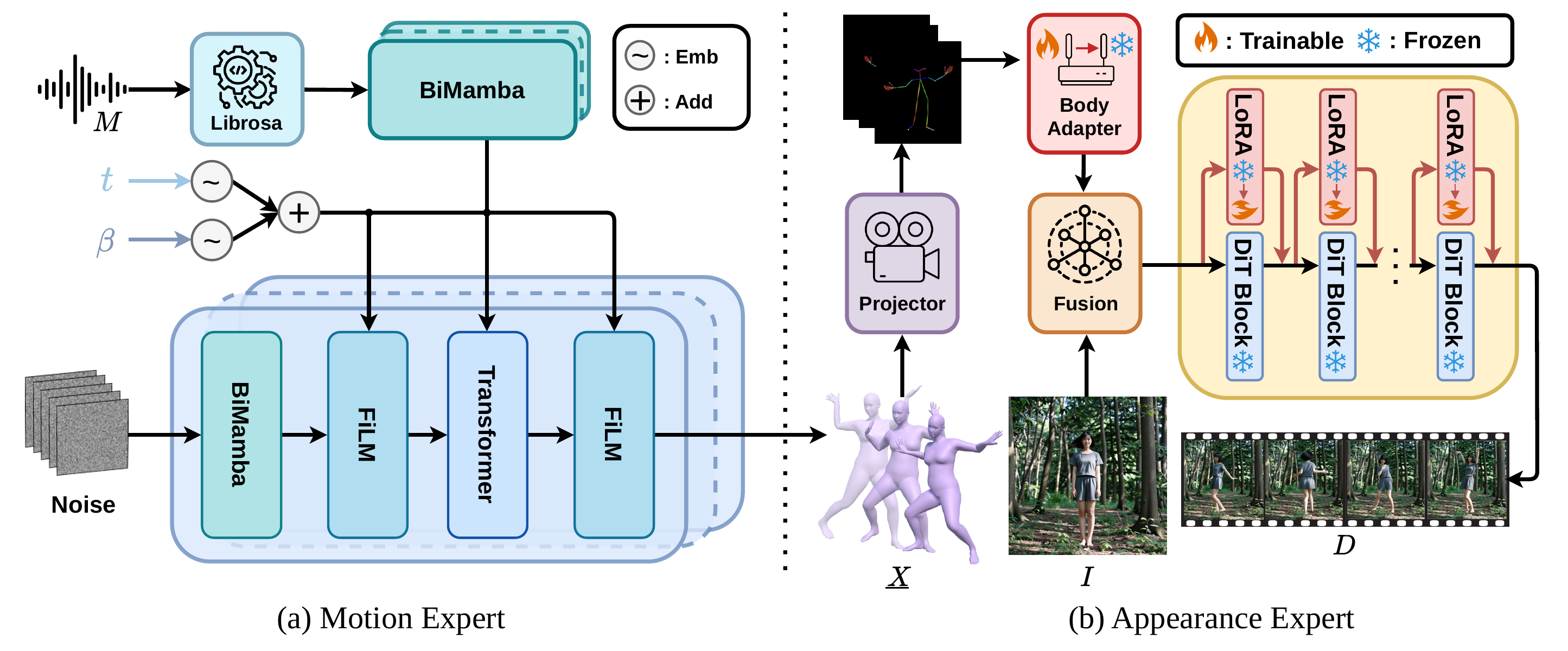}
    % \vspace{-0.2in}
    \caption{Overview of \textbf{MACE-Dance}.Leveraging the cascaded Mixture-of-Experts (MoE) design, the Motion Expert generates kinematically plausible and artistically expressive 3D motion $X$ conditioned on the music $M$, and the Appearance Expert then animates the reference image $I$ with the 3D motion $X$, yielding the dance video $D$ that exhibits spatiotemporally coherent appearance. Thanks to the Guidance-Free Training (GFT) strategy, $\beta \in [0, 1]$ can serve as a controllable knob that governs the diversity of the generated motion.}
    \label{fig: overview}
    % \vspace{-0.17in}
\end{figure*}

Given a music $M \in R^{T \times C_m}$ and reference image $I \in R^{H \times W \times 3}$, our objective is to synthesize the corresponding dance videos $D \in R^{T \times H \times W \times 3}$ with high-quality visual appearance and human motion. Overall, \textbf{MACE-Dance} is with cascaded mixture-of-experts (MoE), as shown in Fig. ~\ref{fig: overview}. The Motion Expert (ME) transfers music sequence $M$ into 3D motion sequence $X\in R^{T \times C_x}$, enforcing kinematic plausibility and artistic expressiveness. The Appearance Expert (AE) utilizes the above 3D motion sequence $X$ and reference image $I$ to drive video synthesis, preserving visual identity with spatiotemporal coherence. This task decoupling significantly reduces the complexity of learning a direct music-to-video mapping by isolating motion semantics from visual appearance. Moreover, the explicit 3D motion representation suppresses spurious cross-modal correlations and provides an interpretable intermediate interface for robust and controllable video synthesis.

Unlike prior works~\cite{chen2025x,tang2025spatial} that adopt 2D keypoints as the intermediate representation, we instead use 3D motion as the bridge between the two experts for three reasons. \textbf{(1) Richer spatial fidelity.} 3D motion preserves full-body geometric structure, including global translation and orientation, which is essential for dance phrases with large-amplitude locomotion and complex spatial choreography, whereas 2D projections inevitably discard depth and global movement information. \textbf{(2) Cleaner supervision.} 3D representation disentangles pose from camera viewpoint and subject-specific appearance, providing a more stable and generalizable signal for learning the music-to-motion correspondence, while 2D keypoints are entangled with perspective and body proportions. \textbf{(3) Better robustness.} 3D motion is inherently more robust to self-occlusion and viewpoint variation, whereas 2D poses often suffer from missing joints, depth ambiguity, and inconsistent observations.
Additionally, we adopt SMPL~\cite{loper2023smpl} as the representation of the 3D motion sequence $X$ for two reasons. \textbf{(1) Prior focus on body motion.} Most existing 3D dance generation methods primarily model body-level motion rather than detailed hand articulation. In our setting, body-level motion alone is sufficient to produce strong visual results, as also evidenced by our demo videos. \textbf{(2) Extensibility.} Our framework can be readily extended to richer motion representations, such as SMPL-X, when suitable data become available.

% Formally:

% \begin{equation}
%     X = ME(M), D = AE(X, I). 
% \end{equation}

\subsection{Motion Expert}
\subsubsection{Generative Strategy.}
\paragraph{Diffusion.} DDPM~\cite{ho2020denoising} defines diffusion as a Markov noising process with latents $\{z_t\}_{t=0}^T$ that follow a forward noising process $q(z_t|x)$, where $x \sim p(x)$ is drawn from the 3D dance data distribution. The forward noising process is defined as:
\begin{equation}
q(z_t | x) \sim \mathcal{N}(\sqrt{\bar{\alpha}_t} x, (1 - \bar{\alpha}_t) I),
\end{equation}
where $\bar{\alpha}_t \in (0,1)$ are constants which follow a monotonically decreasing schedule such that when $\bar{\alpha}_t$ approaches 0. Timestep $T$ are commonly set to 1000, and $z_T \sim \mathcal{N}(0, I)$. With paired music conditioning $c$, we can reverse the forward diffusion process by learning to estimate $\hat{x}_\theta(z_t, t, c) \approx x$ with model parameters $\theta$ for all $t$. We can optimize $\theta$ by the naive reconstruction loss in Diffusion Model~\cite{ho2020denoising}:
\begin{equation}
\mathcal{L}_{\text{DM}} = \mathbb{E}\big[\| \hat{x}_\theta(z_t, t, c) - x  \|_2^2 \big].
\end{equation}

\paragraph{Guidance-Free Training.} 
Conventional classifier-free guidance (CFG~\cite{ho2022classifier}) modifies the sampling distribution only at inference time by combining conditional and unconditional predictions, which can introduce distribution mismatch and insufficient optimization toward the guided target distribution. In contrast, Guidance-Free Training (GFT~\cite{chen2025visual}) retains the same maximum-likelihood training objective as CFG but adopts a different parameterization that enables a single model to implicitly represent temperature-controlled sampling behavior during training, thereby mitigating distribution mismatch and yielding more stable and consistent high-fidelity generation. Accordingly, we establish $x_\beta$ as the new optimization target for our model $\theta$:
\begin{equation}
x_\beta = \beta \hat{x}_\theta(z_t, t, c, \beta) + (1 - \beta) \mathbf{sg}[\hat{x}_\theta(z_t, t, \emptyset, 1)]
\end{equation}
where $\emptyset$ denotes the unconditional setting, and $\mathbf{sg}$ represents the stop-gradient operation. $\beta$ serves as a temperature parameter that is also provided to the model $\theta$ as an additional conditioning input. During training, $\beta$ and $t$ are sampled randomly from $U(0, 1)$ and the integer set $\{0, 1, \dots, T\}$, respectively. Moreover, we further apply the reconstruction loss, 3D joint loss, velocity loss, foot contact loss, to enhance physical plausibility and aesthetic expressiveness:
\begin{equation}
\begin{aligned}
\mathcal{L}_{\text{rec}} &= \mathbb{E}\big[\| x_\beta - x  \|_2^2 \big], \\
\mathcal{L}_{\text{joint}} &= \mathbb{E}\big[\| FK(x_\beta) - FK(x)  \|_2^2 \big], \\
\mathcal{L}_{\text{vel}} &= \mathbb{E}\big[ \| FK(x_\beta)' - FK(x)'  \|_2^2 \big] \\
\mathcal{L}_{\text{foot}} &= \mathbb{E}\big[\| FK(x_\beta)' \cdot \hat{\mathbf{b}}  \|_2^2 \big], \\
\end{aligned}
\end{equation}
where $FK(\cdot)$ denotes the forward kinematic function that converts joint angles into joint positions,  and $\hat{\mathbf{b}}$ is the model's own prediction of the binary foot contact label's portion of the pose. Our overall training loss $\mathcal{L}$ is the weighted sum of the above losses, where the weights $\lambda$ were chosen to balance the magnitudes of the losses:
\begin{equation}
\mathcal{L} = 
\lambda_{\text{rec}} \mathcal{L}_{\text{rec}} + \lambda_{\text{joint}} \mathcal{L}_{\text{joint}} + \lambda_{\text{vel}} \mathcal{L}_{\text{vel}} + \lambda_{\text{foot}} \mathcal{L}_{\text{foot}}.
\end{equation}

\paragraph{Inference.}
At each of the denoising timesteps $t$, Motion Expert predicts the denoised sample and noises it back to timestep $t-1$: $\hat{z}_{t-1} \sim q(\hat{x}_\theta(z_t, t, c, \beta), t-1)$, terminating when it reaches $t = 0$. We utilize Denoising Diffusion Implicit Models (DDIM~\cite{song2020denoising}) to accelerate the sampling procedure. Values of $\beta$ near 0 favor high fidelity, while values near 1 favor high diversity. Thus, $\beta$ can also be regarded as a control signal, and we set its value to 0.75. Notably, GFT inherently achieves theoretically double the generation efficiency compared to conventional CFG, as it only requires a single conditional computation per step, eliminating the need for simultaneous conditional and unconditional predictions.

\subsubsection{Model Architecture}
\paragraph{Overview.}
Motion Expert adopts a BiMamba–Transformer hybrid backbone, thereby enabling the generation of temporally coherent and musically aligned dance motions. BiMamba captures intra-modal local dependencies in music or dance, while the Transformer models cross-modal global context. As shown in Fig. ~\ref{fig: overview}, the architecture details are as follows:
Firstly, our model conditions the generator on the Librosa~\cite{mcfee2015librosa}-extracted music features from $M$ as~\cite{li2021ai}, which are then processed by an $L_m$‑layer BiMamba to capture intra‑modal temporal dynamics. Secondly, the diffusion time step $t$ and temperature parameter $\beta$ are encoded as sinusoidal embeddings and fused by element-wise addition to yield a $t$-$\beta$ embedding used throughout the generator. Third, the dance generator consists of $L_d$ stacked blocks. In each block: (1) the current state $z_t$ is first passed through a BiMamba to model intra-modal local dependencies; (2) FiLM ~\cite{perez2018film} is applied to modulate the features with the fused $t$-$\beta$ embedding; (3) a Transformer performs cross-modal attention over the music encoding to integrate global musical context, and subsequently passes the result through a feed-forward network; and (4) a second FiLM~\cite{perez2018film} further reinforces the $t$-$\beta$ conditioning. Finally, the generator outputs the 3D motion sequence $\hat{x}_\theta(z_t, t, c, \beta)$ (i.e. $X$ in Sec. 3.1 Overview), represented as SMPL~\cite{loper2023smpl} parameters.
Owing to this architecture, the Motion Expert generates the entire sequence in a non-autoregressive manner during inference, not only improving generation efficiency but also avoiding the exposure-bias problem in autoregressive~\cite{yang2025megadance} and inpainting-based~\cite{tseng2023edge} methods.

\paragraph{Intra-Modal Local-Dependency.}
While the Transformer excels at temporal modeling, it is inherently position-invariant and captures sequence order only through positional encodings~\cite{vaswani2017attention}, which limits its deep understanding of local dependencies. In contrast, music-to-dance generation demands strong local continuity between movements. Owing to its inherent sequential inductive bias, Mamba~\cite{gu2023mamba} has demonstrated strong performance in modeling fine-grained local dependencies ~\cite{xu2024mambatalk,fu2024mambagesture}.
Moreover, Bidirectional Mamba processes inputs in both forward and backward directions, enabling wider representations and deeper understanding of music and dance.
Specifically, the Selective State Space Model (Mamba) integrates a selection mechanism and a scan module (S6)~\cite{gu2023mamba} to dynamically emphasize salient input segments for efficient sequence modeling. Unlike traditional SSMs with time-invariant parameters, Mamba generates input-dependent $\bar{A}_t, \bar{B}_t, C_t$ through fully connected layers, enhancing generalization. For each time step $t$, the input $x_t$, hidden state $h_t$, and output $y_t$ evolve as:
\begin{equation}
h_t = \bar{A}_t h_{t-1} + \bar{B}_t x_t,\quad y_t = C_t h_t,
\end{equation}
where $\bar{A}_t, \bar{B}_t, C_t$ are dynamically updated, and the state transitions become:
\begin{equation}
\bar{A} = \exp(\Delta A),\quad
\bar{B} = (\Delta A)^{-1}(\exp(\Delta A)-I)\cdot \Delta B,
\end{equation}
where $\Delta$ is the discretization step size, $A$ is the continuous-time state transition matrix, $B$ is the input projection matrix, and $C$ is the output projection matrix.

\paragraph{Cross-Modal Global-Context.}
While BiMamba~\cite{gu2023mamba} excels at capturing local dependencies, it is less effective at modeling cross-modal global interactions. Thus, we employ a Transformer~\cite{vaswani2017attention} module after BiMamba in each denoising block, which is crucial for aligning the overall dance structure with long-term musical phrasing. This block consists of a cross-attention layer followed by a feed-forward network (FFN). In the cross-attention layer, motion features serve as queries, while music features provide keys and values:
\begin{equation}
\text{Attention} = \text{softmax}\left(\frac{Q_d \cdot K_m ^T}{\sqrt{C}}\right)V_m.
\end{equation}

In this way, the two components play complementary roles: BiMamba stabilizes short-range intra-modal dynamics, while the Transformer injects cross-modal global musical context to align the generated motion with the overall rhythm and phrase structure.

\begin{figure*}[t]
    \centering
    \includegraphics[width=0.9\linewidth]{figs/exp.pdf}
    % \vspace{-0.07in} 
    \caption{Qualitative comparison with SOTAs across reference image domains (real-person vs. anime-character) and music genres (Eastern Folk vs. Popping) in the music-driven dance video generation task.}
    \label{fig: exp}
    % \vspace{-0.1in}
\end{figure*}

\subsection{Appearance Expert}
Wan-Animate~\cite{cheng2025wan} has recently garnered substantial attention in both industry and academia. However, it is designed for general-purpose motion synthesis; direct transfer to dance video generation is suboptimal due to the domain gap and the richer spatiotemporal complexity of dance—namely intricate whole-body coordination and dynamic camera choreography. Accordingly, the Appearance Expert adopts a decoupled Kinematic–Aesthetic fine-tuning strategy to achieve high-fidelity appearance synthesis for dance videos. Specifically, the Kinematic Stage fine-tunes only the Body Adapter while freezing the remaining components, whereas the Aesthetic Stage fine-tunes only the LoRA parameters while keeping the rest of the network fixed.

\paragraph{Model Architecture.}
As illustrated in Fig. ~\ref{fig: overview}, the architecture of our Appearance Expert is built upon the Wan-Animate~\cite{cheng2025wan}, which takes a reference image $I$ for appearance and a 3D motion sequence $X$ for motion guidance. 
The motion sequence $X$ is first projected to 2D keypoints, which are then encoded by a Body Adapter to yield motion features. 
These features are subsequently fused with the latent extracted from the reference image $I$. 
The resulting latent is processed by a backbone of stacked DiT blocks, where lightweight LoRA adapters are integrated into each block. 
The facial processing pipeline remains identical to that of Wan-Animate and is therefore omitted for clarity.

\paragraph{Projector.}
We introduce a 3D-to-2D Motion Projector to convert the SMPL sequence generated by the Motion Expert into the 2D pose format required by Wan-Animate. For each frame, we first transform the SMPL parameters into a 3D mesh and render it with \texttt{pyrender} under a fixed frontal-view camera, then apply \texttt{ViTPose}~\cite{xu2022vitpose} to extract the corresponding 2D keypoints. In this way, the projector preserves the benefits of 3D motion modeling while enabling seamless integration with the downstream Appearance Expert.

\paragraph{Kinematic Stage.}
In dance, body pose is paramount. The original Wan‑Animate prioritizes facial cues, allocating a dedicated cross‑attention branch to the face while fusing body signals only via additive injection. We therefore strengthen kinematic conditioning by fine‑tuning the Body Adapter in the Kinematic Stage to reweight and calibrate body features across scales, thereby enforcing motion adherence without altering the backbone. We intentionally do not introduce an additional body cross‑attention branch because (i) it disturbs the pretrained inductive bias and can compete with the facial cross‑attention, causing feature entanglement, (ii) it adds substantial memory/latency overhead and training instability on long, fast dance sequences.

\paragraph{Aesthetic Stage.}
To refine visual quality without disturbing motion control, we freeze the kinematic pathways and attach lightweight LoRA adapters to the attention (query/key/value/output) and feed-forward projections in each DiT block of Wan‑Animate. These rank‑r adapters enable parameter‑efficient specialization toward dance‑specific aesthetics—sharpening textures (skin, hair, fabric), stabilizing clothing and accessories, and handling rich camera choreography (pans, zooms, handheld motion)—while preserving pretrained content priors. Specifically, LoRA is an effective technique for adapting large pre-trained models to down-streaming tasks with few training-able parameters. To achieve this goal, LoRA introduces a low-rank decomposition-based method to the model's weight matrix, enabling efficient adaptation to new tasks while maintaining the model's original capabilities. Given the weight matrix $W_0 \in \mathbb{R}^{m \times n}$ of the original pre-trained model, LoRA~\cite{hu2022lora} uses two low-rank matrices $A \in \mathbb{R}^{m \times r}$ ($r \ll m$) and $B \in \mathbb{R}^{r \times n}$ ($r \ll n$) to shift the trained distribution according to the new data training. Thanks to the low-rank matrix in $A$ and $B$, LoRA updates the model more efficiently than the full rank matrix and shows comparable results with full training. Formally, the new weight matrix $W$ can be represented as:
\begin{equation}
    W = W_0 + \Delta W = W_0 + AB.
\end{equation}

\begin{table*}[t]
\centering
\renewcommand{\arraystretch}{1.2}
\setlength{\tabcolsep}{5pt}
\caption{Quantitative comparison with SOTAs on the MA-Data dataset in Music-Driven Dance Video Generation task.}
% \vspace{-0.15in}
\label{tab: dance video comparison}
\scriptsize % 在这里设置字号
\resizebox{0.95\linewidth}{!}{ % 调整了宽度以更好地适应减少的列
\begin{tabular}{l|cccccc|ccccc} % 在方法列和两大分类之间加竖线以区分
\toprule
 & \multicolumn{6}{c|}{\textbf{Appearance}} & \multicolumn{5}{c}{\textbf{Motion}} \\
\cmidrule(r){2-7} \cmidrule(l){8-12} % 在分类下方绘制部分横线
 & IQ$\uparrow$ & AQ$\uparrow$ & SC$\uparrow$ & BC$\uparrow$ & MS$\uparrow$ & TF$\uparrow$
 & FID$_{k}$$\downarrow$ & FID$_{g}$$\downarrow$ & DIV$_{k}$$\uparrow$ & DIV$_{g}$$\uparrow$ & BAS$\uparrow$ \\ 
\midrule
Ground Truth          
& 67.12	& 53.51	& 91.86	& 92.97	& 98.20	& 96.88 
& --     & --     & 9.24 & 5.31 & 0.526 \\

Hallo2~\cite{cui2024hallo2} [ICLR'25]
& 62.64 & \underline{50.79} & 92.48 & 93.84 & 98.30 & 96.56
& \underline{16.55} & \underline{1.29} & 8.11  & 5.47 & \underline{0.505} \\

WAN-S2V~\cite{gao2025wan} [Arxiv'25]
& 64.10 & 50.20 & 92.30 & 93.40 & 98.20 & 96.70
& 18.90 & 1.45 & 7.60 & 5.44 & 0.485 \\

Echomimic-V3~\cite{meng2025echomimicv3} [AAAI'26]
& 63.20 & 49.00 & 91.90 & 93.10 & 98.05 & 96.40
& 19.60 & 1.32 & 7.20 & 4.60 & 0.460 \\

EDGE~\cite{li2023finedance} [CVPR'23]
& 63.05 & 49.70 & 91.79 & 93.30 & \textbf{98.64} & \textbf{97.10}
& 21.77 & 1.39 & \underline{9.08} & \underline{5.74} & 0.498 \\

Lodge~\cite{li2024lodge} [CVPR'24] 
& 63.69 & 49.22 & 91.67 & 92.98 & 98.46 & \underline{97.05}
& 18.73 &  1.49 & 8.87  &  5.71 &  0.474 \\

MEGA~\cite{yang2025megadance} [NeurIPS'25]
& \textbf{66.14} & 49.89 & \underline{92.95} & \underline{94.13} & 97.45 & 96.32
& 18.98 & 1.65 & 8.78 & 5.59 & 0.513 \\

\midrule % 添加一条分割线，突出我们的方法
MACE-Dance
& \underline{65.35} & \textbf{51.79} & \textbf{93.97} & \textbf{94.57} & \underline{98.46} & \textbf{97.10}  % <- 6列结果已留空
& \textbf{16.46} & \textbf{0.28} & \textbf{9.74} & \textbf{6.34} & \textbf{0.523} \\
\bottomrule
\end{tabular}
}
% \vspace{-0.1in}
\end{table*}

\begin{table}[t]
\centering
\renewcommand{\arraystretch}{1.2}
\setlength{\tabcolsep}{5pt}
\caption{Quantitative comparison with SOTAs on the FineDance dataset in Music-Driven 3D Dance Generation task.}
% \vspace{-0.15in}
\label{tab: 3D Dance Comparison}
\scriptsize % 在这里设置字号
\resizebox{0.95\linewidth}{!}{ % 宽度可以根据需要调整
\begin{tabular}{l|ccccccc} % 只在第一列后保留竖线
\toprule
 & FID$_{k}$$\downarrow$ & FID$_{g}$$\downarrow$ & FSR$\downarrow$
 & DIV$_{k}$$\uparrow$ & DIV$_{g}$$\uparrow$
 & BAS$\uparrow$ 
 & FPS$\uparrow$ \\ 

\midrule
Ground Truth          
& --     & --     & 0.216
& 9.94 & 7.54 & 0.201
& -- \\

FACT~\cite{li2021ai} [ICCV'21]  
& 113.38 & 97.05 & 0.284
& 3.36  & 6.37 & 0.183
& 29 \\

MNET~\cite{kim2022brand} [CVPR'22]    
& 104.71 & 90.31 & 0.394
& 3.12  & 6.14 & 0.186
& 26 \\

Bailando~\cite{siyao2022bailando} [CVPR'22]      
& 82.81 & 28.17 & \underline{0.188}
& 7.74  & 6.25 & 0.202
& 188 \\

EDGE~\cite{tseng2023edge} [CVPR'23]      
& 94.34 & 50.38 & 0.200
& \underline{8.13}  & \underline{6.45} & 0.212
& 119 \\

Lodge~\cite{li2024lodge} [CVPR'24]      
& \underline{50.00} &  35.52 & \textbf{0.028}
& 5.67  &  4.96 &  \underline{0.226}
& 224 \\

MEGA~\cite{yang2025megadance} [NeurIPS'25]
& \underline{50.00} & \textbf{13.02} & 0.243
& 6.23 & 6.27 & \underline{0.226}
& \underline{238} \\

\hline
GFT $\rightarrow$ CFG
& 25.54 & 36.31 & 0.295
& \textbf{11.57} & 7.03 & 0.223
& 475 \\

BiMamba $\rightarrow$ Mamba
& 65.10 & 51.74 & 0.35
& 9.59 & 7.66 & 0.224
& \textbf{1044} \\

BiMamba $\rightarrow$ Transformer
& 104.93 & 114.42 & 0.12
& 7.90 & 4.22 & 0.266
& 683 \\

\textbf{Motion Expert} (Full) 
& \textbf{17.83} & \underline{25.09} & 0.210
& \textbf{10.30} & \textbf{8.09} & \textbf{0.229}
& \textbf{770} \\

\bottomrule
\end{tabular}
}
% \vspace{-0.2in}
\end{table}

\section{Experiment}
\subsection{Dataset}
To support music-driven dance video generation task, we curate a large-scale dance video dataset, named MA-Data. It comprises ~70k clips of 5–10 seconds each (totaling 116 hours), and spans over 20 distinct dance genres, such as Jazz, Latin, Eastern Folk. MA-Data consists of two complementary sources. \textbf{(1) 3D-rendered data} (motion-centric). This subset is derived from FineDance~\cite{li2023finedance}, the largest 3D dance dataset recorded by professional dancers, and emphasizes professional dance motion rather than visual appearance. Specifically, we first retarget the motion sequence to a character model, then render front-view videos from the 3D character, and extract random 5–10 s segments with a sliding-window strategy for data augmentation, yielding ~20k clips (~28 hours).  
\textbf{(2) In-the-wild internet data} (appearance-centric). This subset is curated from high-engagement creators on platforms such as TikTok and YouTube, emphasizing visual appearance; motions typically prioritize entertainment value over technical rigor. As raw crawls include many samples misaligned with our task, we apply a multi-stage cleaning pipeline: (i) perform shot boundary detection with TransNet V2~\cite{soucek2024transnet}, segment accordingly, and discard segments shorter than 5 s; (ii) remove near-static clips using an optical-flow magnitude threshold; (iii) enforce a single-performer constraint via ViTPose~\cite{xu2022vitpose} by discarding clips that contain multiple people or exhibit little to no human motion; and (iv) split long videos into 5–10 s clips with a sliding window and random offsets. The final set comprises ~50k clips (~88 hours). Finally, we collect an additional 200 5-second clips to construct the test set, with high engagement on TikTok and across multiple dance genres.

\subsection{Evaluation}
The key challenges of music-driven dance video generation are~\cite{yang2024beatdance,zhang2025robust}: (1) generating dance \textbf{motions} that are kinematically plausible while artistically expressive; and (2) achieving high-fidelity visual \textbf{appearance} with strong spatiotemporal consistency. Inspired by this, we introduce a motion–appearance evaluation protocol. 
\textbf{(1) Motion dimension}. We extract 2D keypoint sequences using ViTPose~\cite{xu2022vitpose} from the dance videos and evaluate from a Human-Kinematics perspective. To evaluate the fidelity and diversity, we report FID and DIV across two feature spaces~\cite{li2021ai, siyao2022bailando}: (1) kinetic (k), capturing motion dynamics, and (2) geometric (g), encoding spatial joint relations. To measure music–motion synchronization, we utilize the Beat Alignment Score (BAS)~\cite{li2021ai, li2024lodge}.
\textbf{(2) Appearance dimension}. Inspired by ~\cite{chen2025finger,li2025ld,ling2025vmbench}, we adopt VBench~\cite{huang2024vbench}—a widely used benchmark in video generation—and select a set of dance-specific metrics. Our evaluation includes imaging quality (IQ), aesthetic quality (AQ), subject consistency (SC), background consistency (BC), motion smoothness (MS), temporal flickering (TF).

\begin{table}[t]
\centering
\renewcommand{\arraystretch}{1.2}
\setlength{\tabcolsep}{5pt}
\caption{Quantitative comparison with SOTAs on the MA-Data dataset in Pose-Driven Image Animation task.}
% \vspace{-0.15in}
\label{tab: image animation comparison}
\tiny
\resizebox{0.95\linewidth}{!}{ % 我稍微增大了宽度以容纳新列，您可以根据需要调整
\begin{tabular}{l|cccc}
\toprule
 & FVD$\downarrow$ & SSIM$\uparrow$ & LPIPS$\downarrow$ & PSNR$\uparrow$ \\
\midrule

Animate-Anyone~\cite{hu2024animate} [CVPR'24]   
& 515.26 & 0.648 & 0.091 & 19.65 \\

Magic-Animate~\cite{xu2024magicanimate} [CVPR'24]
& 1032.06 & 0.311 & 0.207 & 14.00 \\

Wan-Animate~\cite{cheng2025wan} [Arxiv'25]
& \underline{332.82} & \underline{0.707} & \underline{0.078} & \underline{21.11} \\

\hline
w/o. Kinematic stage
& 328.91 & 0.596 & 0.107 & 18.69 \\
w/o. Aesthetic stage
& 445.93 & 0.563 & 0.121 & 17.89 \\
Appearance Expert
& \textbf{274.94} & \textbf{0.739} & \textbf{0.066} & \textbf{22.40} \\

\bottomrule
\end{tabular}
}
% \vspace{-0.2in}
\end{table}

\begin{figure*}[!t]
    \centering
    \includegraphics[width=0.95\linewidth]{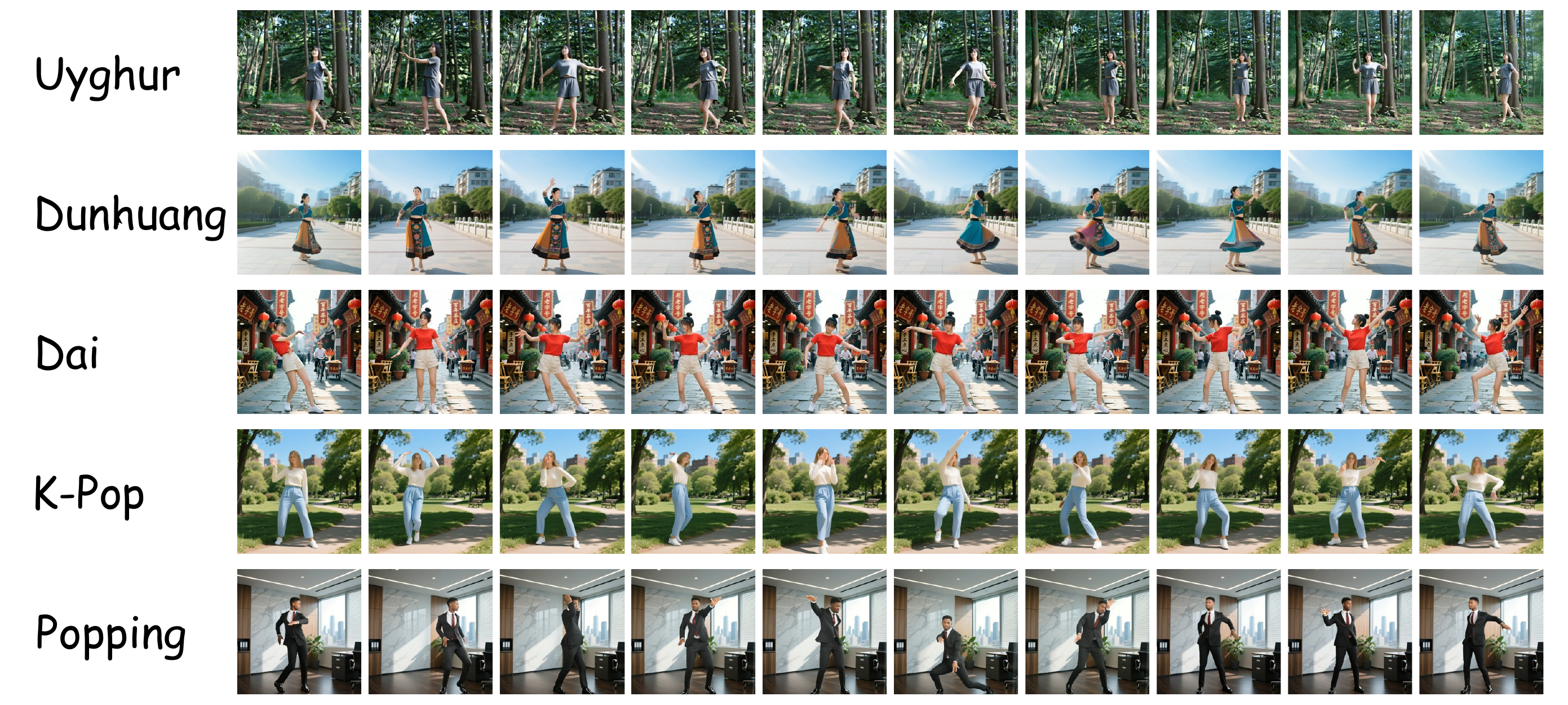}
    % \vspace{-0.15in}
    \caption{\textbf{MACE-Dance} generates high-quality dance videos across diverse dance genres.}
     % \vspace{-0.15in}
    \label{fig: supp_genre}
\end{figure*}

\subsection{Comparison}
\subsubsection{Music-Driven Dance Video Generation.} 
As there is currently no open-source implementation for Music-Driven Dance Video Generation, we compare \textbf{MACE-Dance} against two baseline families on the MA-Data dataset: (1) 3D dance generation methods pipelined with Wan-Animate, including EDGE~\cite{tseng2023edge}, Lodge~\cite{li2024lodge}, and MEGA~\cite{yang2025megadance}; (2) General human-motion video generation methods. We perform inference using the pretrained EchoMimic-V3~\cite{meng2025echomimicv3} and WAN-S2V~\cite{gao2025wan} models, which supports general human motion generation. In addition, we adapt the Hallo2~\cite{cui2024hallo2} by replacing facial masks with full-body masks, and then fine-tune it on the MA-Data dataset.
As shown in Tab. ~\ref{tab: dance video comparison}, our proposed \textbf{MACE-Dance} demonstrates state-of-the-art (SOTA) performance in both appearance and motion quality. 
Specifically, for the motion aspect, \textbf{MACE-Dance} achieves the best results across all metrics ($FID_k=16.46, FID_g=0.28, DIV_k=9.74, DIV_g=6.34, BAS=0.523$); for the appearance aspect, it also attains best performance on most metrics, with scores of $IQ=65.35, AQ=51.79, SC=93.97, BC=94.57, MS=98.46,$ and $TF=97.10$.
By effectively decoupling the task into 3D Dance Generation and Pose-Driven Image Animation, and leveraging the strong performance of the Motion Expert and Appearance Expert on their respective sub-tasks, \textbf{MACE-Dance} delivers exceptional generation quality. \textit{Note: User Study can be found in the supplementary material Sec. 2.}

\subsubsection{Music-Driven 3D Dance Generation.}
Music-Driven 3D Dance Generation is a canonical task and determines the motion quality of \textbf{MACE-Dance}. We compare the Motion Expert against FACT~\cite{li2021ai}, MNET~\cite{kim2022brand}, Bailando~\cite{siyao2022bailando}, EDGE~\cite{tseng2023edge}, Lodge~\cite{li2024lodge}, and MEGA~\cite{yang2025megadance} on the FineDance dataset with metrics following ~\cite{li2024lodge}. As shown in Tab. ~\ref{tab: 3D Dance Comparison}, the Motion Expert attains overall state-of-the-art (SOTA) performance. Specifically, it achieves the best $FID_k = 17.83$ and a competitive $FID_g = 25.09$, indicating high fidelity; the best $DIV_k = 10.30$ and $DIV_g = 8.09$, indicating strong diversity; a competitive FSR, supporting physical plausibility; the best $BAS = 0.229$, demonstrating superior audio-motion synchronization; and a substantially higher $FPS = 770$, evidencing excellent generation efficiency. These advances primarily stem from: (1) adopting a Diffusion Model with a BiMamba-Transformer hybrid architecture, enabling high-quality long motion sequences in a non-autoregressive manner; and (2) a Guidance-Free Training (GFT) strategy that improves generation quality without requiring dual-pass inference (conditioned + unconditioned). \textit{Note: Qualitative Comparison can be found in the supplementary material Sec. 3.2.}

\subsubsection{Pose-Driven Image Animation.}
Pose-Driven Image Animation is likewise a canonical task and governs the appearance quality of \textbf{MACE-Dance}. We compare the Appearance Expert against Animate-Anyone~\cite{hu2024animate}, Magic-Animate~\cite{xu2024magicanimate}, and Wan-Animate~\cite{cheng2025wan} on the MA-Data dataset with metrics following ~\cite{cheng2025wan}. The Appearance Expert achieves state-of-the-art (SOTA) performance on all metrics (Tab. \ref{tab: image animation comparison}, $FVD=274.94$, $SSIM=0.739$, $LPIPS=0.066$, $PSNR=22.40$). Its strong video synthesis quality primarily benefits from Wan-Animate (Baseline)'s powerful cross-modal understanding and the Kinematic-Aesthetic decoupled fine-tuning strategy. \textit{Note: Qualitative Comparison can be found in the supplementary material Sec. 3.3.}

\begin{figure*}[t]
\centering

% ================= Left column =================
\begin{minipage}[t]{0.48\linewidth}
\vspace*{0pt}      % ⭐ 强制左列视觉顶部
\centering

\includegraphics[width=0.95\linewidth]{figs/abl.pdf}
% \vspace{-0.15in}
\captionof{figure}{Ablation for the model architecture of Motion Expert.}
\label{fig: abl me}

\vspace{0.08in}    % ✅ 用正的间距控制

\includegraphics[width=\linewidth]{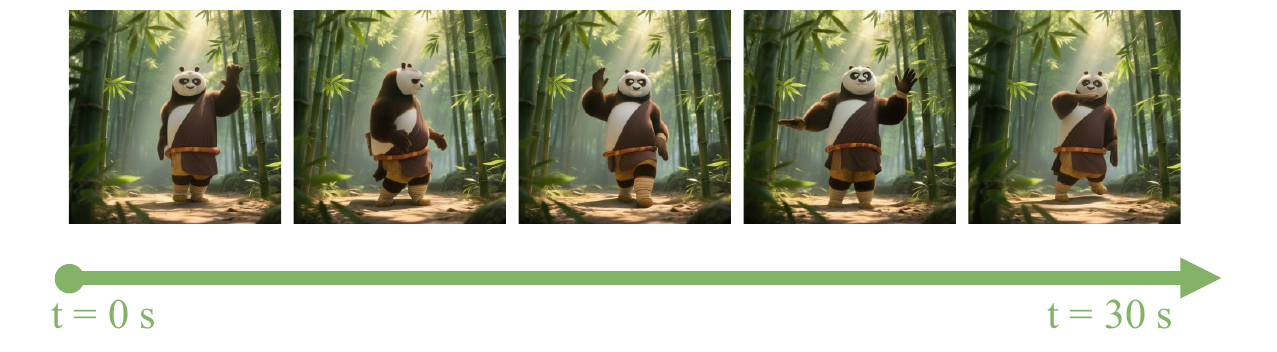}
% \vspace{-0.15in}
\captionof{figure}{\textbf{MACE-Dance} produces coherent long-sequence dance videos.}
\label{fig: supp_longdance}

\end{minipage}
\hfill
% ================= Right column =================
\begin{minipage}[t]{0.48\linewidth}
\vspace*{0pt}      % ⭐ 强制右列视觉顶部
\centering

\includegraphics[width=0.95\linewidth]{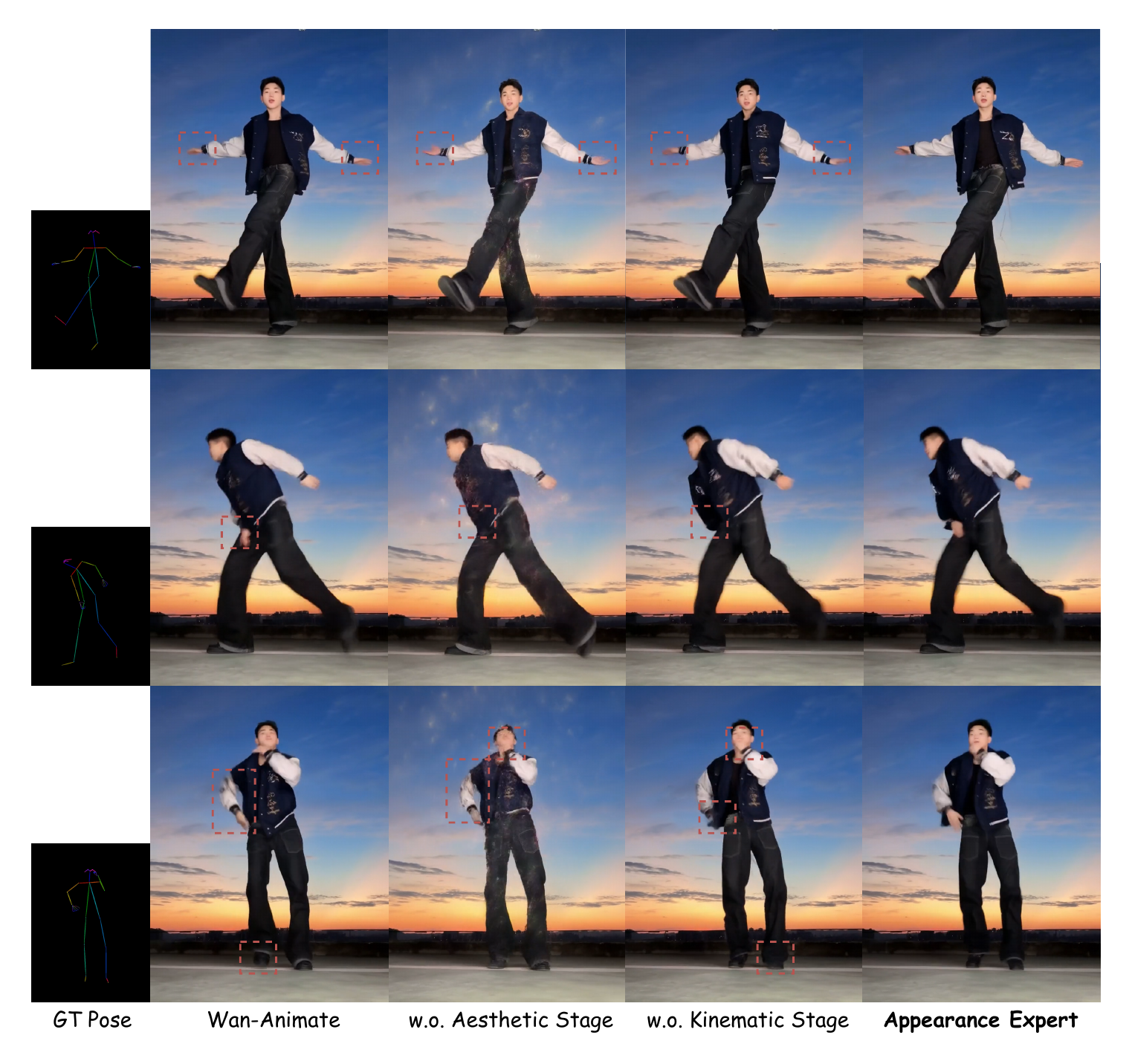}
% \vspace{-0.15in}
\captionof{figure}{Ablation for the Appearance Expert.}
\label{fig: abl ae}

\end{minipage}
\end{figure*}

\subsection{Qualitative Analysis}
\subsubsection{Effect Comparison.} 
We also present a qualitative comparison against other methods across reference-image domains (real person vs. anime character) and music genres (elegant and rhythmically rich Eastern Folk vs. powerful and funk-inspired Popping) for the music-driven dance video generation task. As shown in Fig. ~\ref{fig: exp}, Hallo2 exhibits significant blurring in human details and noticeable artifacts; EDGE often shows abrupt motion discontinuities; Lodge frequently produces abnormal movements that violate physical plausibility; MEGA, WAN-S2V, and Echomimic-V3 often produce overly simple and repetitive motions, limiting expressiveness. In contrast, videos generated by \textbf{MACE-Dance} not only present kinematically plausible and artistically expressive human motion, but also maintain spatiotemporally coherent visual appearance.

\subsubsection{Cross-Genre Generation.} 
Moreover, \textbf{MACE-Dance} generalizes effectively across dance genres as shown in Fig. ~\ref{fig: supp_genre}, producing distinct genre-specific motion signatures, including (1) Uyghur dance exhibits light, continuous upper-body rotations with expressive arm trajectories; (2) Dunhuang motion features stable lower-body stances and elegant, circular arm patterns; (3) Dai style emphasizes soft, flowing wrist and elbow movements; (4) K-Pop example demonstrates crisp transitions, symmetrical poses, and rhythm-driven gestures; (5) Popping is characterized by sharp isolations and staccato movements, reflecting its percussive movement vocabulary.

\subsubsection{Long-Sequence Generation.}
Additionally, a complete music track typically lasts 30 seconds to 5 minutes, making long-sequence generation crucial for practical dance video synthesis~\cite{feng2025narrlv}. To mitigate motion drift or visual degradation in long-sequence generation, \textbf{MACE-Dance} incorporates dedicated designs in both stages: (1) a BiMamba–Transformer hybrid in the Motion Expert for drift-free long motion synthesis, and (2) pose-driven relay rendering with identity anchoring in the Appearance Expert. As shown in Fig. ~\ref{fig: supp_longdance}, \textbf{MACE-Dance} produces coherent long-sequence dance video.

% In summary, \textbf{MACE-Dance} demonstrates superior performance in qualitative evaluations.

\subsection{Ablation Study}
\subsubsection{Motion Expert}
Given that the BiMamba–Transformer hybrid architecture and the Guidance-Free Training (GFT) strategy are central to the Motion Expert’s performance, we ablate them individually to assess their effects, as shown in Tab. ~\ref{tab: 3D Dance Comparison} and Fig. ~\ref{fig: abl me}. 
\textbf{(1) Model Architecture.} Replacing BiMamba with Mamba removes bidirectional context, weakening temporal understanding. Quantitatively, although generation efficiency improves, all dance-quality metrics degrade, indicating this is not a worthwhile trade-off. Qualitatively, the generated dances tend to resort to simple, common movements, diminishing the model’s artistic expressiveness. Replacing BiMamba with Transformer deprives the model of its ability to generate dance in a non-autoregressive manner, owing to self-attention’s scale-dependent positional extrapolation. Quantitatively, most metrics drop to unacceptable levels. Qualitatively, the model collapses to in-place side-to-side jitter, i.e., a poor local optimum. This also explains why BAS and FSR increase instead—these gains come at the cost of severely compromised dance quality.
\textbf{(2) Generative Strategy.} Replacing GFT with naive classifier-free guidance (CFG) leads to a modest decline across most metrics. Notably, our generation efficiency improves by approximately 1.62$\times$, because inference requires only a single conditional setting. \textit{Note: The effect of $\beta$ in GFT can be found in the supplementary material Sec. 4.2.}

\subsubsection{Appearance Expert}
Since the two-stage fine-tuning strategy is central to our Appearance Expert, we evaluate the contribution of each stage via ablation, as summarized in Tab. ~\ref{tab: image animation comparison} and Fig. ~\ref{fig: abl ae}. \textbf{(1) Kinematic Stage.} We fine-tune the Body Adapter while freezing all other components. Removing this stage leads to a modest decline across all metrics quantitatively and noticeable kinematic errors and motion blur qualitatively, indicating its effectiveness for ensuring human kinematic plausibility in video. \textbf{(2) Aesthetic Stage.} We fine-tune LoRA parameters in each DiT block. Omitting this stage causes a substantial degradation across all metrics quantitatively and obvious ghosting artifact qualitatively, underscoring its critical role to preserve video aesthetic. Finally, Appearance Expert also demonstrates the superior performance over the Wan-Animate (Baseline), which validates the overall effectiveness of the proposed Kinematic-Aesthetic fine-tuning strategy.

\subsubsection{Motion Representation (2D vs. 3D)}
Most pose-driven image animation methods rely on 2D keypoints, which naturally motivates using 2D poses as the intermediate motion representation. In contrast, \textbf{MACE-Dance} adopts 3D motion as its intermediate representation. To validate this design, we compare 2D and 3D representations at both the motion-generation level and the final video-generation level. Specifically, for the motion level, we train the same Motion Expert on FineDance with either 2D or 3D motion targets. For the video level, we further render final videos on MA-Data using either 2D pose sequences directly or 3D motion projected to 2D via our projector. As shown in Tab.~\ref{tab: motion representation}, the 3D-based representation consistently outperforms the 2D-based one across both settings. On FineDance, 3D achieves better fidelity, diversity, and synchronization, indicating that it provides more stable and physically consistent supervision for music-to-motion learning. On MA-Data, 3D further yields clearly better subject consistency, visual fidelity, and beat alignment in the final rendered videos, showing that the advantages of 3D are preserved after the downstream animation stage. These results demonstrate that 3D motion serves as a more reliable intermediate interface than 2D pose for both controllable dance generation and high-quality video synthesis.

\begin{table}[t]
\centering
\renewcommand{\arraystretch}{1.2}
\setlength{\tabcolsep}{5pt}
\caption{Comparison of 2D and 3D motion representations.}
% \vspace{-0.1in}
\label{tab: motion representation}
% \scriptsize
\resizebox{0.95\linewidth}{!}{
\begin{tabular}{l|ccccc|cccc}
\toprule
& \multicolumn{5}{c|}{Kinematic} 
& \multicolumn{4}{c}{Appearance} \\
\cmidrule(r){2-6} \cmidrule(l){7-10}
 
& FID$_k$$\downarrow$ & FID$_g$$\downarrow$ & DIV$_k$$\uparrow$ & DIV$_g$$\uparrow$ & BAS$\uparrow$
& AQ$\uparrow$ & SC$\uparrow$ & FID$\downarrow$ & BAS$\uparrow$ \\
\midrule
2D & 22.8 & 8.6 & 6.12 & 5.24 & 0.527 & \textbf{51.86} & 91.84 & 23.73 & 0.496 \\
3D & \textbf{19.5} & \textbf{4.1} & \textbf{8.87} & \textbf{5.92} & \textbf{0.543} & 51.79 & \textbf{93.97} & \textbf{16.46} & \textbf{0.523} \\
\bottomrule
\end{tabular}
}
% \vspace{-0.12in}
\end{table}

\begin{table}[t]
    \centering
    \caption{Cross-composition analysis of the two experts on MA-Data.}
    \label{tab:expert_contribution}
    \resizebox{0.65\linewidth}{!}{
    \begin{tabular}{lcccc}
        \toprule
        Method & AQ$\uparrow$ & SC$\uparrow$ & FID$\downarrow$ & BAS$\uparrow$ \\
        \midrule
        w/o.ME & 50.21 & 92.10 & 20.84 & 0.499 \\
        w/o.AE & 50.36 & 91.42 & 17.92 & 0.519 \\
        Ours   & \textbf{51.79} & \textbf{93.97} & \textbf{16.46} & \textbf{0.523} \\
        \bottomrule
    \end{tabular}
    }
\end{table}

\subsubsection{Role of Each Expert}
To further analyze the role of each expert in \textbf{MACE-Dance}, we conduct an additional cross-composition study on MA-Data by replacing one expert at a time with its corresponding baseline counterpart. Specifically, \textit{w/o.ME} denotes the variant that uses the baseline \textbf{Motion Expert} (\texttt{EDGE~\cite{tseng2023edge}}) together with our \textbf{Appearance Expert}, while \textit{w/o.AE} denotes the variant that uses our \textbf{Motion Expert} together with the baseline \textbf{Appearance Expert} (\texttt{WAN-Animate~\cite{cheng2025wan}}). As shown in Tab.~\ref{tab:expert_contribution}, the full \textbf{MACE-Dance} consistently achieves the best performance across all evaluated metrics, confirming that both experts contribute positively to the final music-driven dance video generation quality. More specifically, replacing our \textbf{Appearance Expert} with the baseline model leads to clear degradation in appearance-related metrics such as \textit{AQ} and \textit{SC}, while replacing our \textbf{Motion Expert} results in a more noticeable drop in the motion-related metric \textit{BAS}. These observations suggest that the two experts play complementary roles: the \textbf{Motion Expert} mainly strengthens music-motion alignment and body dynamics, whereas the \textbf{Appearance Expert} further improves visual quality, temporal coherence, and identity consistency in the rendered videos.

\subsection{Comparison with Video Foundation Models}
\begin{figure}[t]
    \centering
    \includegraphics[width=0.9\linewidth]{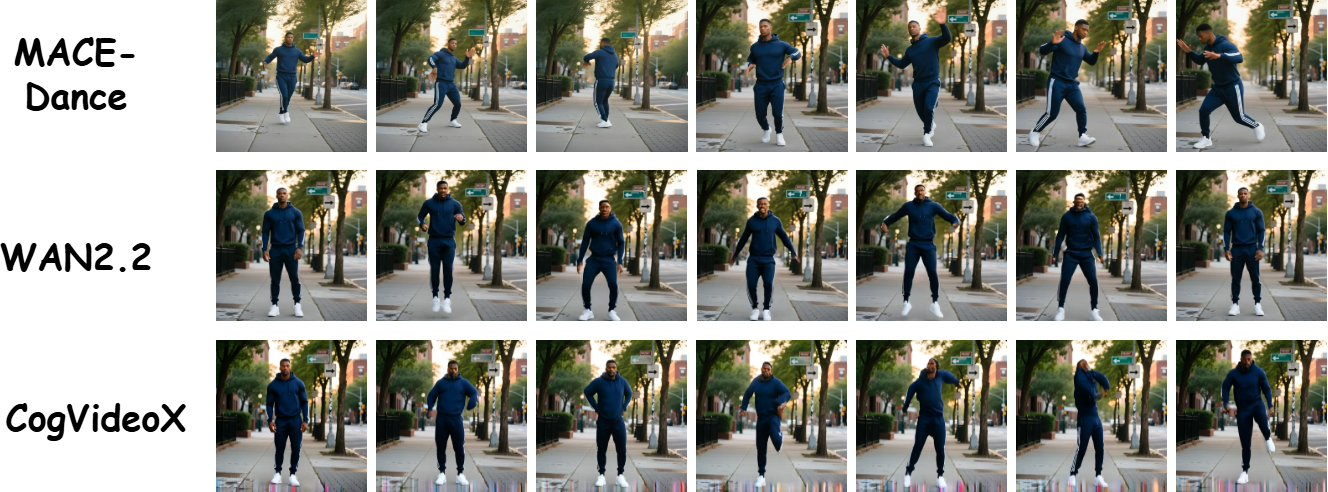}
    % \vspace{-0.2in}
    \caption{Comparison with general-purpose video foundation models.}
    \label{fig: ti2v}
    % \vspace{-0.17in}
\end{figure}

We further compare MACE-Dance with general-purpose video foundation models, including CogVideoX1.5-5B~\cite{yang2024cogvideox} and WAN2.2-5B~\cite{wan2025wan}, to examine how a structured motion-to-appearance pipeline performs against recent large-scale video generation models. Although these models demonstrate strong generation ability in broad video domains, they are not specifically designed for music-driven dance video generation, where accurate modeling of beat, rhythm, and body-motion coherence is particularly important. As shown in Table~\ref{tab:vfm_compare}, MACE-Dance achieves the best overall performance on \textit{SC}, \textit{FID}, and \textit{BAS}, indicating stronger music-motion alignment, better visual quality, and more consistent identity preservation. Although WAN2.2-5B attains a slightly higher \textit{AQ} score, it underperforms our method on the other three metrics. Qualitatively, CogVideoX1.5-5B tends to produce weaker and slower dance motions with noticeable blur, while WAN2.2-5B generates larger motion amplitudes but often suffers from temporal identity inconsistency, as shown in Fig.~\ref{fig: ti2v}. Overall, these results support the effectiveness of explicitly decomposing music-driven dance video generation into a Motion Expert and an Appearance Expert.

\begin{table}[t]
    \centering
    \caption{Comparison with general-purpose video foundation models.}
    \label{tab:vfm_compare}
    \resizebox{0.7\linewidth}{!}{
    \begin{tabular}{lcccc}
        \toprule
        Method & AQ$\uparrow$ & SC$\uparrow$ & FID$\downarrow$ & BAS$\uparrow$ \\
        \midrule
        CogVideoX1.5-5B & 50.38 & 89.92 & 22.47 & 0.477 \\
        WAN2.2-5B       & \textbf{53.22} & 90.77 & 17.53 & 0.452 \\
        Ours            & 51.79 & \textbf{93.97} & \textbf{16.46} & \textbf{0.523} \\
        \bottomrule
    \end{tabular}
    }
\end{table}

\section{Conclusion}
In conclusion, we present \textbf{MACE-Dance}, a music-driven dance video generation framework with cascaded Mixture-of-Experts (MoE). The Motion Expert enforces kinematic plausibility and artistic expressiveness, while the Appearance Expert preserves visual identity with spatiotemporal coherence. Specifically, the Motion Expert adopts Diffusion Model with a BiMamba–Transformer hybrid backbone and Guidance-Free Training strategy, while the Appearance Expert adopts a decoupled Kinematic–Aesthetic fine-tuning strategy. To better benchmark this task, we curate a large-scale dataset, and design a motion–appearance evaluation protocol. Extensive experiments demonstrate the superiority of \textbf{MACE-Dance} and of its Motion and Appearance Experts. For future work, we plan to extend MACE-Dance with textual descriptions to enable more interactive and flexible dance generation, and improve system-level efficiency to support low-latency authoring and real-time user feedback.

% \section{Acknowledge}
% This work was supported in part by the National Nature Science Foundation of China under Grants  62436010, 72572090, 62572474 and 62172421, and in part by Tsinghua University School of Economics and Management Research Grant.
\begin{acks}
This work was supported in part by the National Natural Science Foundation of China under Grants  62436010, 72572090, 62572474 and 62172421, and in part by Tsinghua University School of Economics and Management Research Grant.
\end{acks}

\balance
\bibliographystyle{ACM-Reference-Format}
\bibliography{MACEDance}
\clearpage

% \begin{figure*}[!t]
%     \centering
%     \includegraphics[width=\linewidth]{figs/supp_cmp3.pdf}
%     \caption{Qualitative comparison with SOTAs on hip-hop music in the music-driven dance video generation task.}
%     \label{fig: case}
% \end{figure*}

\section{Implementation Details}
MACE-Dance is a music-driven dance video generation framework with cascaded Mixture-of-Experts (MoE), decoupling this task into music-to-3D motion generation task (Motion Expert) and pose-driven image animation task (Appearance Expert). Additionally, due to the specific data requirements of each expert, the Motion Expert is trained exclusively on the 3D-rendered, motion-centric subset of the data, while the Appearance Expert is trained on the entire MA-Data dataset. We will introduce them in turn.

\subsection{Motion Expert}
For Motion Expert, we adopt the Diffusion Model with BiMamba-Transformer hybrid architecture and Guidance-Free Training (GFT) strategy on the FineDance datasets. For training setup, we adopt the Adam optimizer with a learning rate $4 \times 10^{-4}$, weight decay $0.02$. The model is trained for 4000 epochs with a batch size of 128 using the \text{Accelerate} library for distributed training on 8 NVIDIA H20 Tensor Core GPUs. We train on sequences of 240 frames (8s) and perform inference on sequences of 1024 frames (34.13s). EMA (decay $0.9999$) is applied to stabilize training, and checkpoints are periodically saved for evaluation (50 epochs). We combine multiple objectives:  reconstruction loss ($\lambda_{rec}$=0.636), 3D joint position loss ($\lambda_{joint}$=0.636), velocity loss ($\lambda_{vel}$=2.964) and foot contact loss ($\lambda_{foot}$=10.942). For model architecture, The conditional processing part contains 2 layers of BiMamba with Genre-Gate, and the vector generation part includes 8 layers of BiMamba-Transformer-based block. Each Mamba unit sets state 16, convolutional kernel size $4$, and expansion factor $2$, and latent dimension $512$; each Transformer block utilizes 4 attention heads, a feed-forward network dimension of 1024, a dropout rate of 0.1, and the GELU activation function. We set temperature parameter $\beta$ in GFT 0.75 during inference.

\subsection{Appearance Expert}
For Appearance Expert, we adopt the Kinematic-Aesthetic decoupled fine-tuning strategy on the MA-Data Dataset. 
In the \textbf{Kinematic Stage}, we exclusively fine-tune the Body Adapter to strengthen kinematic conditioning while freezing the entire DiT backbone and VAE. Training is conducted on NVIDIA H20 Tensor Core GPUs. We employ the Adam optimizer with a learning rate of $1 \times 10^{-5}$ and a batch size of 128. This stage is trained for 50k iterations using the standard simple diffusion noise prediction loss to ensure strict motion adherence without altering the pre-trained generative prior.
In the \textbf{Aesthetic Stage}, we freeze the kinematic pathways and fine-tune the extended LoRA branches to capture dance-specific visual patterns. We insert low-rank adapters with rank $r=32$ into the query, key, value, and output projections of the attention modules, as well as the feed-forward networks (FFN) within each DiT block. This stage is optimized using Adam with a learning rate of $5 \times 10^{-5}$ for 50k iterations, minimizing the reconstruction loss to refine texture fidelity and spatiotemporal coherence. The training is distributed across 128 NVIDIA H20 GPUs.

\section{User Study}

\begin{figure}[!t]
  \centering
  % \begin{subfigure}{\linewidth}
    % 将这里的 width=\linewidth 改为一个更小的值，比如 0.8\linewidth
    \includegraphics[width=1.0\linewidth]{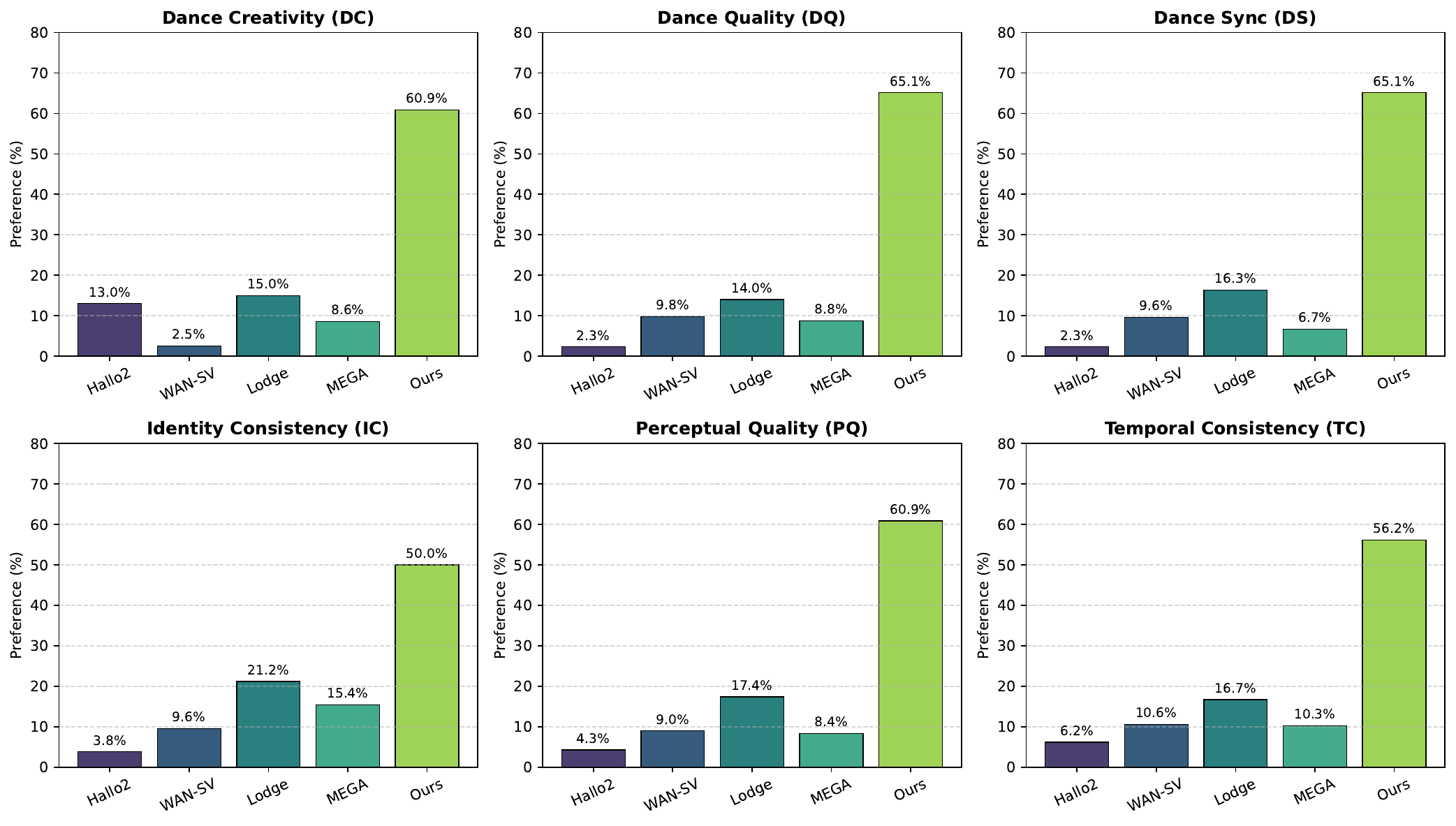}
    
    \caption{\textbf{User study results comparing our method with four baselines.} The bar charts display the percentage of user preferences across six dimensions: Dance Synchronization (DS), Dance Quality (DQ), Dance Creativity (DC), Perceptual Quality (PQ), Temporal Consistency (TC), and Identity Consistency (IC). Our method (Ours) consistently achieves the highest preference rates across all motion and appearance metrics.}
    % \vspace{-15pt}
    \label{fig: userstudy}
  % \end{subfigure}
\end{figure}

\begin{figure*}[!t]
    \centering
    \includegraphics[width=\linewidth]{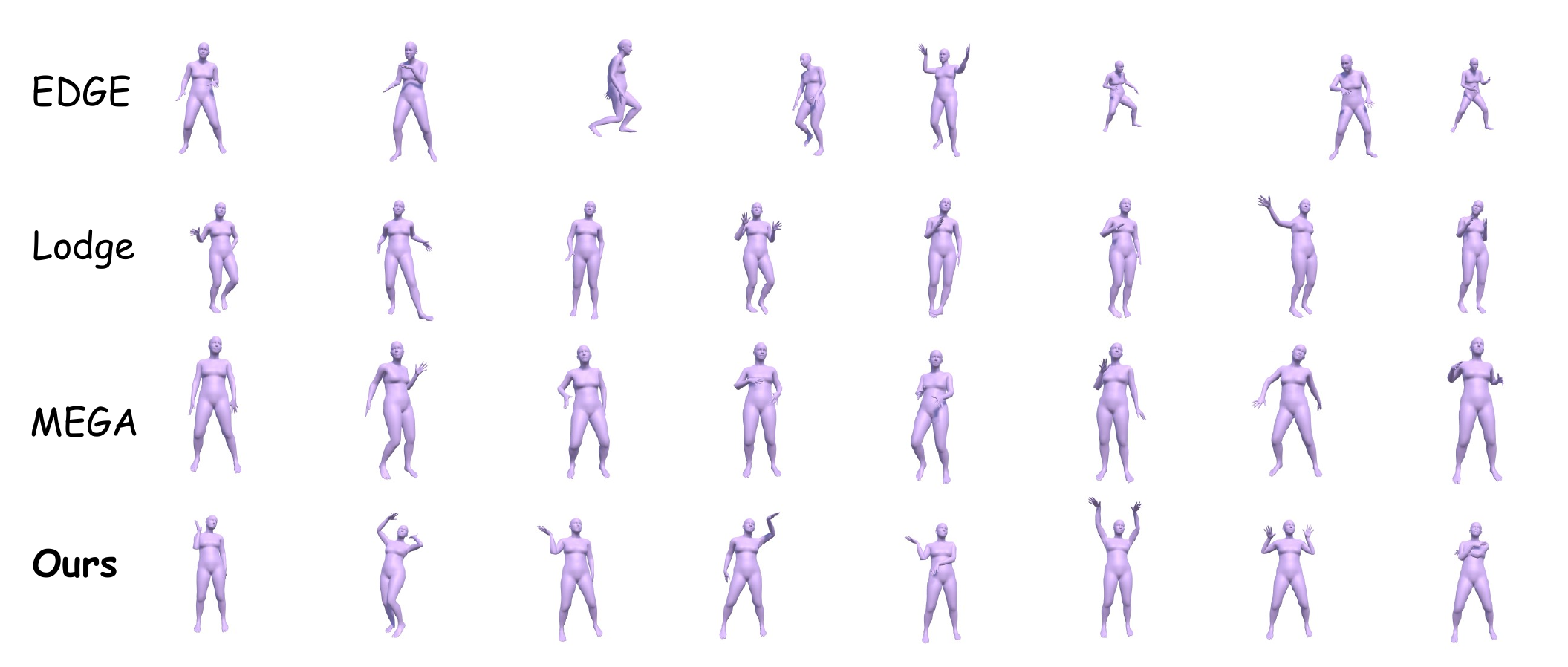}
    \caption{\textbf{Motion Expert} in \textbf{MACE-Dance} can generate high-quality 3D Motion with artistic expressiveness and physical plausibility.}
    \label{fig: supp_3D}
\end{figure*}

\subsection{Experimental Setting}
User feedback is essential for evaluating generated dance movements in the music-to-dance generation task, due to the inherent subjectivity of dance~\cite{legrand2009perceiving}. Following~\cite{yang2025megadance}, we select 30 real-world music segments, each lasting 8 seconds, and generated dance sequences using the models described in main paper Sec. 4.3.1. These sequences are evaluated through a double-blind questionnaire completed by 40 participants with dance backgrounds, including undergraduate and graduate students. Participants are compensated at a rate exceeding the local average hourly wage.

Different from scoring individual videos in isolation, we adopted a \textbf{preference-based ranking} mechanism to capture subtle differences between methods. For each query, participants were presented with generated videos from five distinct methods (our proposed method and four baselines) displayed side-by-side in randomized order. For every test case, participants were asked to select all videos that performed best according to the specific criteria, allowing for multiple selections in cases where several methods exhibited equally superior performance.

The evaluation was conducted across six dimensions, categorized into two key challenges of music-driven dance video generation:
\textbf{(1) Human Motion}, which focuses on kinematic plausibility and artistic expressiveness. This includes: 
\begin{itemize}
    \item \textbf{Dance Synchronization (DS)}: Alignment with rhythm and style.
    \item \textbf{Dance Quality (DQ)}: Physical plausibility and aesthetic expressiveness.
    \item \textbf{Dance Creativity (DC)}: Originality and diversity of the movements.
\end{itemize}
\textbf{(2) Visual Appearance}, which focuses on high-fidelity rendering and spatiotemporal consistency. This includes:
\begin{itemize}
    \item \textbf{Perceptual Quality (PQ)}: Naturalness and overall aesthetic quality.
    \item \textbf{Temporal Consistency (TC)}: Smoothness and consistency of the subject and background over time.
    \item \textbf{Identity Consistency (IC)}: Maintenance of the subject's identity relative to the reference image.
\end{itemize}

\begin{figure*}[!t]
    \centering
    \includegraphics[width=\linewidth]{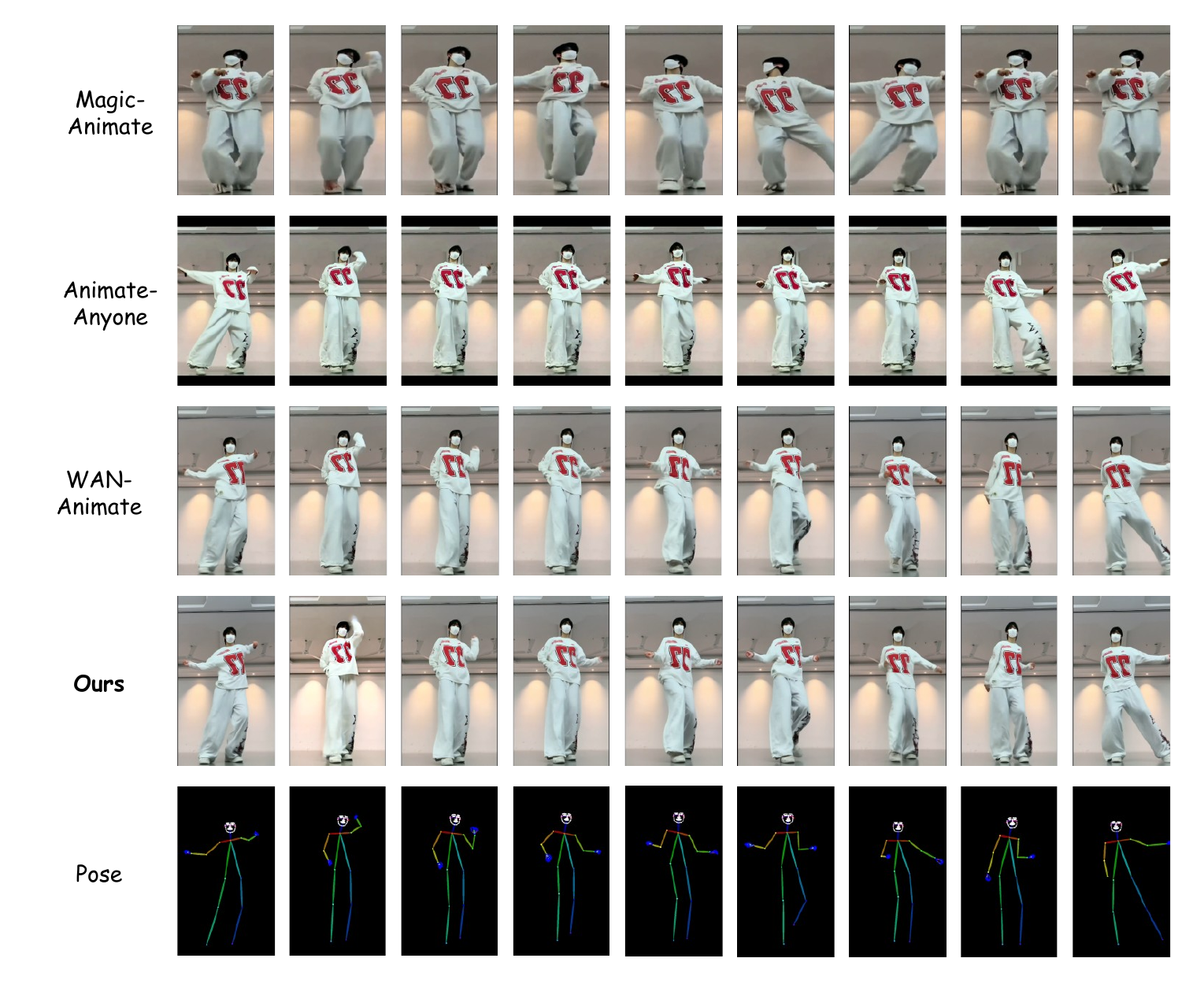}
    \caption{The \textbf{Appearance Expert} in \textbf{MACE-Dance} drives image-based dancing with spatiotemporally coherent appearance.}
    \label{fig: supp_pose}
\end{figure*}

\subsection{Result Analysis}
The quantitative results of the user study are summarized in Fig.~\ref{fig: userstudy}. Our method demonstrates a dominant preference rate across all six evaluated dimensions, significantly outperforming the four baseline methods.

\textbf{(1) Motion Performance.} 
In terms of human motion generation, our method achieves the highest user preference. Specifically, for \textbf{Dance Synchronization (DS)} and \textbf{Dance Quality (DQ)}, our method received over 60\% of the user votes. This indicates that our approach not only aligns dance beats more precisely with the music rhythm but also generates kinematically more plausible and aesthetically pleasing movements compared to competitors. Notably, in \textbf{Dance Creativity (DC)}, our method also leads by a substantial margin, suggesting that our model avoids repetitive patterns and produces more diverse choreographic sequences.

\textbf{(2) Appearance Performance.} 
Regarding visual quality, users overwhelmingly preferred our results. For \textbf{Perceptual Quality (PQ)} and \textbf{Identity Consistency (IC)}, our method secured the vast majority of preferences, validating the effectiveness of our generation pipeline in preserving fine-grained details and subject identity. Furthermore, the high preference rate in \textbf{Temporal Consistency (TC)} demonstrates our model's superior ability to maintain stability across frames, effectively mitigating flickering and temporal artifacts that are common in baseline methods.

Overall, the user study results align with the qualitative visualizations, confirming that our method sets a new state-of-the-art standard in both motion expressiveness and visual fidelity.

\subsection{Evaluation Analysis}
To examine whether the proposed motion–appearance evaluation protocol aligns with human perception, we compare the quantitative results of Tab. 1 in main paper with the User Study outcomes presented in Fig.~\ref{fig: userstudy}. Across all six human-rated dimensions—Dance Creativity (DC), Dance Quality (DQ), Dance Sync (DS), Identity Consistency (IC), Perceptual Quality (PQ), and Temporal Consistency (TC)—\textbf{MACE-Dance} is overwhelmingly preferred by participants, with preference ratios ranging from 50\% to 65.1\%. These human judgments exhibit strong correspondence with our quantitative metrics.

\textbf{(1) Motion Side}. Methods ranked highest by participants on DQ and DS are exactly those achieving superior FID$_k$, FID$_g$, DIV$_k$, DIV$_g$, and BAS scores. In particular, the substantial improvement of \textbf{MACE-Dance} in BAS (0.523), FID$_g$ (0.28), and DIV$_k$ (9.74) is mirrored by its leading human preference in DQ (65.1\%) and DS (65.1\%). This demonstrates that our motion metrics faithfully capture the perceptual qualities that users associate with expressive, synchronized, and natural dance movement.

\textbf{(2) Appearance Side}. The VBench-derived metrics (IQ, AQ, SC, BC, MS, TF) show clear alignment with user ratings in PQ and IC. For example, \textbf{MACE-Dance} achieves the highest scores in SC (93.97), BC (94.57), and TF (97.10), which directly correspond to its large margins in Identity Consistency (50.0\%) and Temporal Consistency (56.2\%) in the user study. Similarly, its strong IQ and AQ scores coincide with the highest PQ rating (60.9\%) among all compared methods.

Taken together, the strong consistency between quantitative metrics and human preference validates the effectiveness of our motion–appearance evaluation protocol. This confirms that the proposed metrics not only provide reliable automatic assessment but also closely reflect human perceptual judgments, making them a meaningful and principled framework for evaluating music-driven dance video generation.

\begin{figure*}[t]
    \centering

    % --- Case 1 ---
    % \begin{subfigure}{\linewidth}
    %     \centering
    %     \includegraphics[width=\linewidth]{figs/supp_cmp3.pdf}
    %     \caption{Case 1}
    % \end{subfigure}

    % --- Case 2 ---
    \begin{subfigure}{0.95\linewidth}
        \centering
        \includegraphics[width=\linewidth]{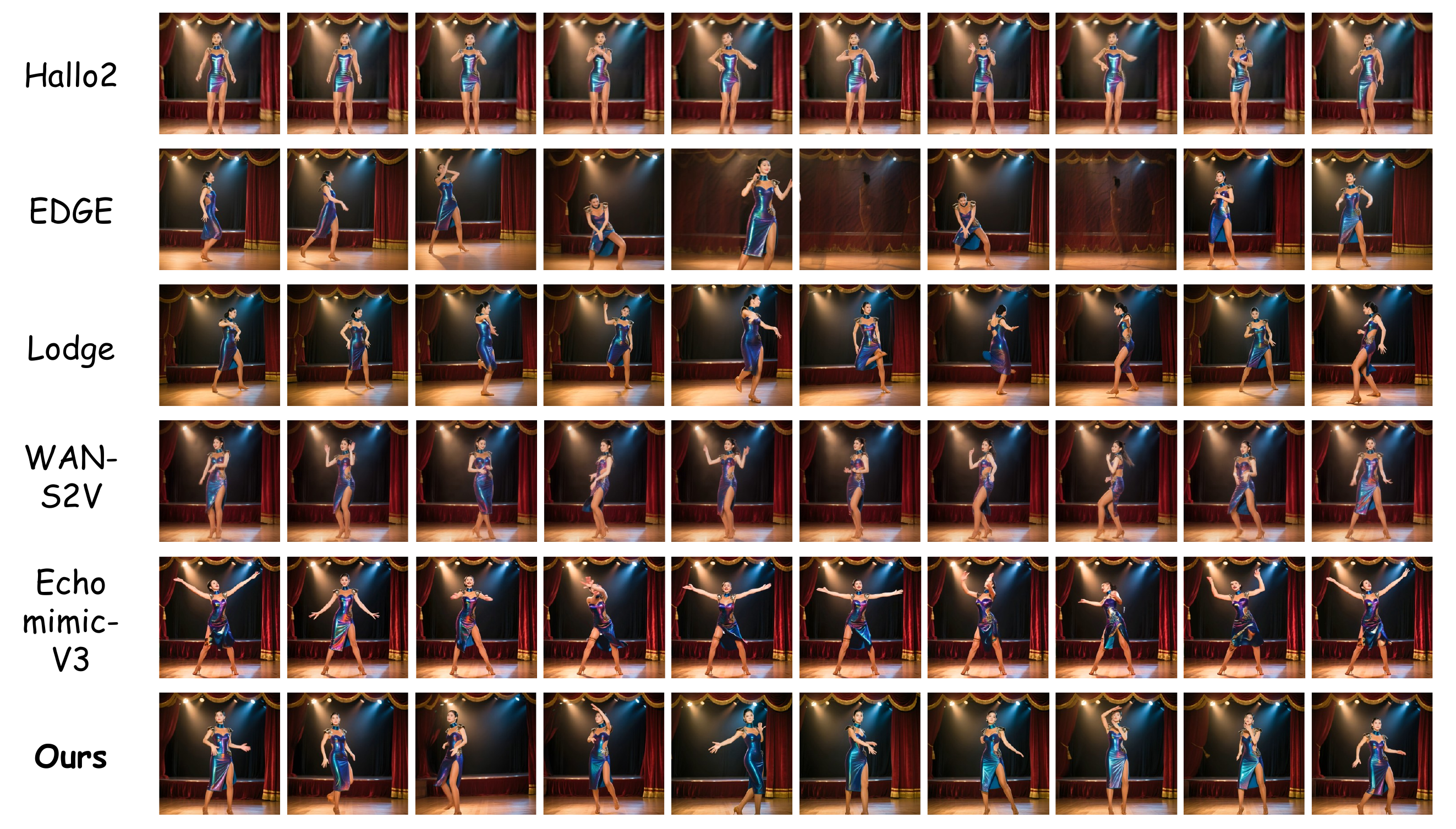}
        \vspace{-0.25in}
        \caption{Case 1}
    \end{subfigure}

    % --- Case 3 ---
    \begin{subfigure}{0.95\linewidth}
        \centering
        \includegraphics[width=\linewidth]{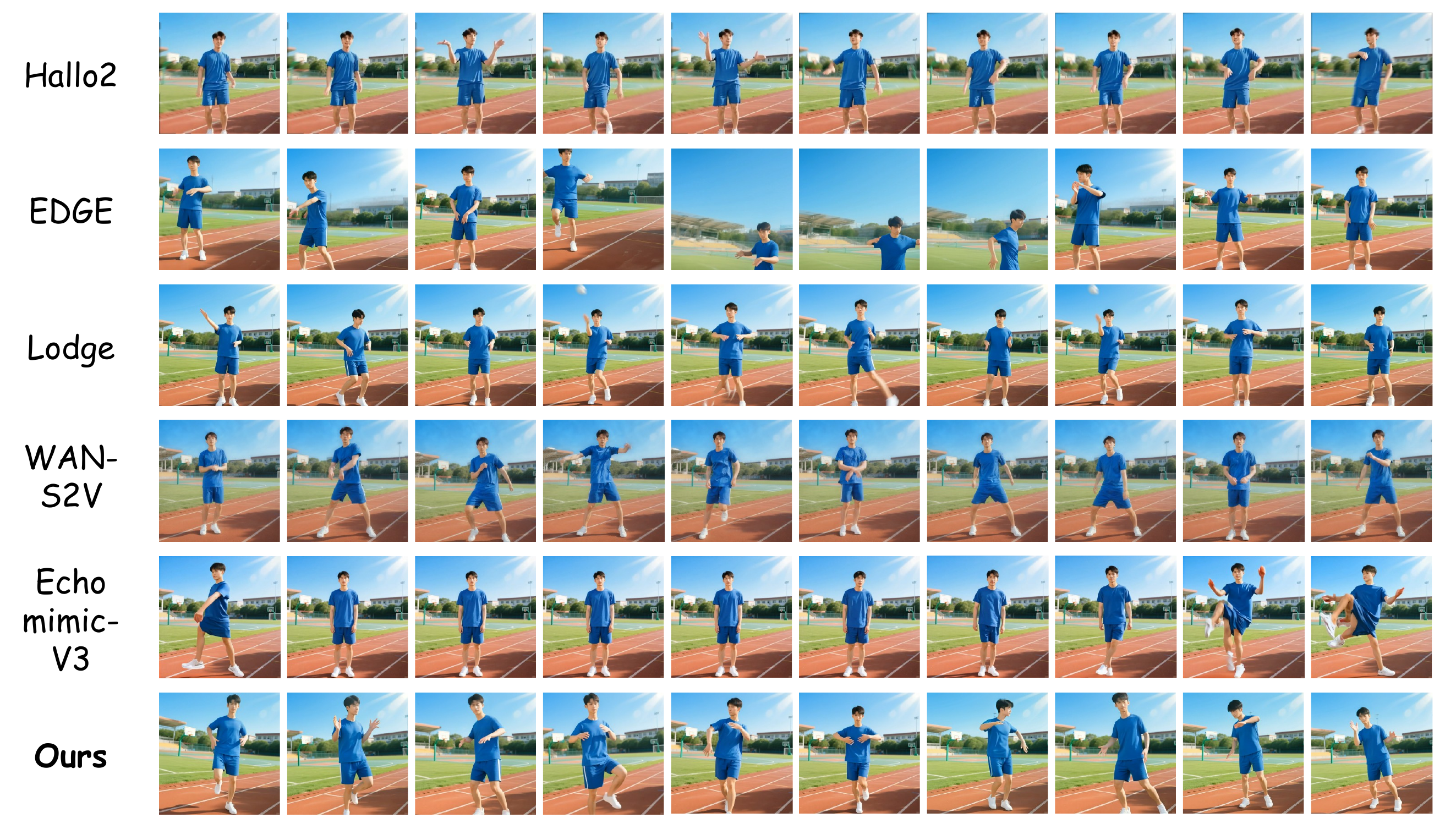}
        \vspace{-0.25in}
        \caption{Case 2}
    \end{subfigure}
    \vspace{-0.1in}
    \caption{More cases of qualitative comparison with SOTAs in music-driven dance video generation task.}
    \label{fig: supp_music-driven dance video generation}
\end{figure*}

\section{Qualitative Analysis}
\subsection{Music-Driven Dance Video Generation}
We further provide additional qualitative comparisons across diverse reference-image domains and music genres to complement the evaluations in the main paper Sec. 4.3.1. As illustrated in Fig.~\ref{fig: supp_music-driven dance video generation}, Hallo2 produces blurred facial regions and introduces substantial background artifacts; EDGE frequently suffers from abrupt motion discontinuities that degrade temporal smoothness; Lodge often yields physically implausible body configurations and irregular motion patterns; and WAN-S2V and Echomimic-V3 tends to generate overly simplified and repetitive motion sequences that lack expressive variety.
In contrast, our method (\textbf{MACE-Dance}) generates videos with kinematically plausible and artistically expressive movements while preserving spatiotemporally coherent appearance across frames. These results further validate the superior qualitative performance of \textbf{MACE-Dance} across a wide range of reference-image inputs and musical styles.

\subsection{Music-Driven 3D Dance Generation}
We also conduct qualitative analyses for the music-driven 3D dance generation task. As shown in Fig. ~\ref{fig: supp_3D}, the observations are consistent with those reported in Sec 3.1 of the Appendix. Specifically, EDGE exhibits abrupt motion discontinuities that compromise temporal smoothness; Lodge often produces physically implausible body configurations and irregular motion patterns; and MEGA tends to generate overly simplified and repetitive motion sequences with limited expressive diversity.
In contrast, our \textbf{Motion Expert} synthesizes 3D motion that is both kinematically plausible and artistically expressive, demonstrating stable dynamics and rich stylistic detail. These results further validate the superiority of the proposed \textbf{Motion Expert} in modeling high-quality, music-driven 3D dance motion.

\subsection{Pose-Driven Image Animation}
We further conduct qualitative comparisons against Magic-Animate, Animate-Anyone, and Wan-Animate on the MA-Data test set. As shown in Fig.~\ref{fig: supp_pose}, existing methods exhibit several limitations when dealing with dance-specific motion patterns. Magic-Animate and Animate-Anyone often produce noticeable spatial distortions and temporal flickering in fast or large-amplitude motions, leading to unstable body shapes and inconsistent textures across frames. Wan-Animate, while stronger in preserving subject identity, still struggles with motion adherence—particularly in rapid limb movements—resulting in lagging body parts and partial pose mismatch.
These qualitative observations highlight the advantage of our two-stage specialization and demonstrate that the proposed \textbf{Appearance Expert} effectively adapts general-purpose image animation models to the unique demands of pose-driven dance video synthesis.

\begin{figure}[t]
    \centering
    \includegraphics[width=0.95\linewidth]{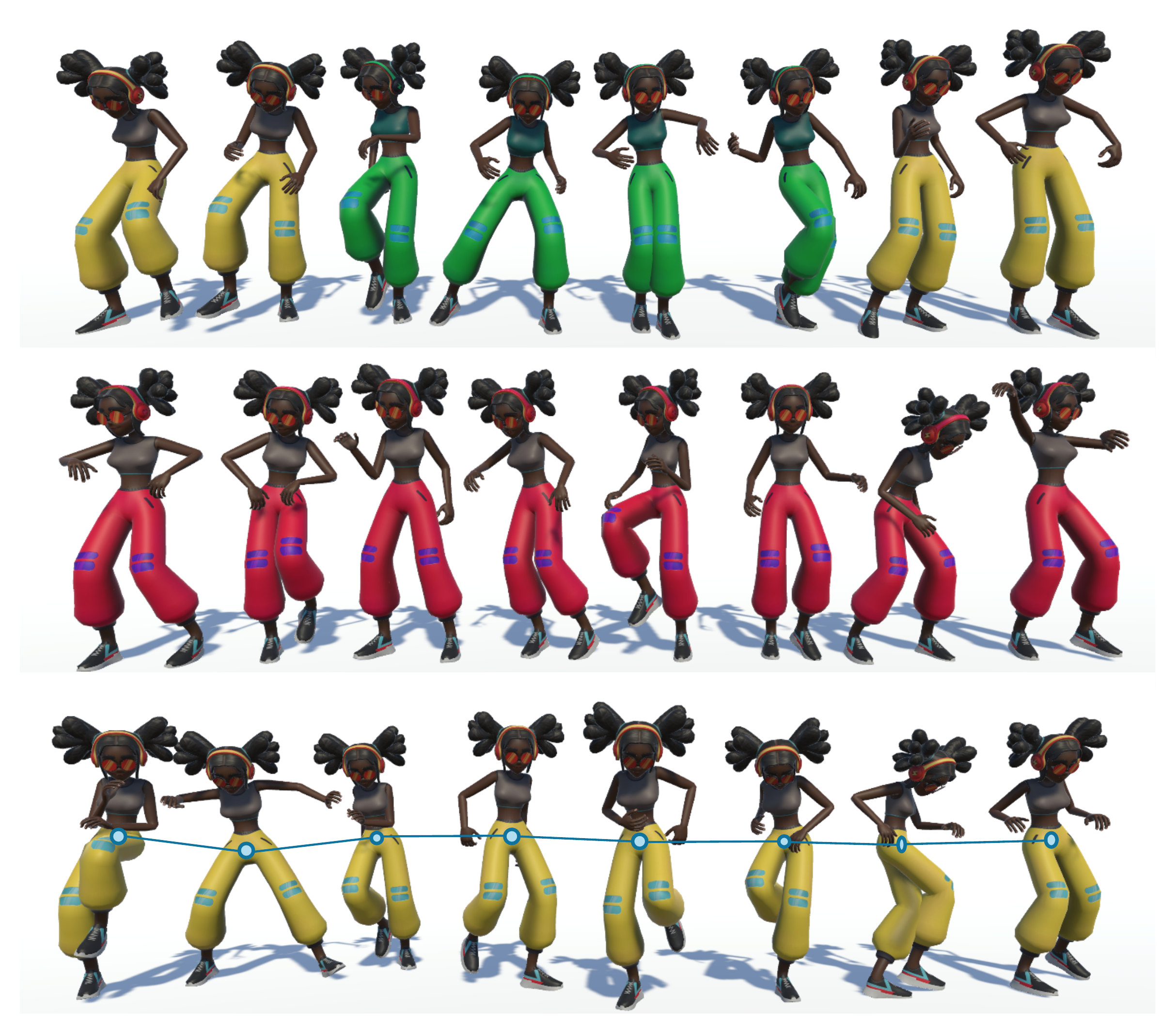}
    \caption{Visualization for motion editing. From top to bottom, the first row shows temporal-level motion editing (yellow indicates the given motion sequence, and green indicates the completed part); the second row shows joint-level motion editing (the upper body indicates the given motion sequence, and the red lower body indicates the completed part); the third row shows trajectory-level completion (given the blue trajectory, the motion sequence is completed accordingly).}
    \label{fig:supp_motion_editing}
\end{figure}

\section{Motion Editing}

Beyond unconditional music-driven dance synthesis, the \textbf{Motion Expert} in \textbf{MACE-Dance} also supports motion editing at inference time through a masked denoising strategy, similar to diffusion-based inpainting. Since the model operates on structured 3D motion sequences rather than pixels, it can preserve user-specified motion constraints while plausibly completing the remaining unknown regions. Formally, let a motion sequence be denoted as $x \in \mathbb{R}^{N \times D}$, where $N$ is the number of frames and $D$ is the motion dimensionality. Given a partial motion constraint $x^{\mathrm{known}}$ together with a binary mask $m \in \{0,1\}^{N \times D}$, where $m_{ij}=1$ indicates that the corresponding element is fixed, we perform masked denoising at each reverse step by replacing the constrained region with the forward-diffused version of the known motion at the same noise level:
\begin{equation}
\tilde{z}_{t-1}
=
m \odot q(x^{\mathrm{known}}, t-1)
+
(1-m) \odot \hat{z}_{t-1},
\label{eq:motion_editing}
\end{equation}
where $\hat{z}_{t-1}$ is the current reverse sample predicted by the model, $q(x^{\mathrm{known}}, t-1)$ denotes the forward diffusion of the known motion to timestep $t-1$, and $\odot$ is element-wise multiplication. In this way, the constrained region remains faithful to the user-provided motion signal, while the unconstrained region is generated by the diffusion prior to ensure temporal smoothness, physical plausibility, and musical coherence. Importantly, this editing mechanism is fully compatible with our DDIM-based inference and requires no additional training.

As illustrated in Fig.~\ref{fig:supp_motion_editing}, this formulation naturally supports three practical editing modes through different mask designs. First, \textbf{temporal inpainting} preserves motion at the beginning and/or end of a sequence and synthesizes the missing middle part, enabling motion in-betweening and smooth transition generation. Second, \textbf{joint-wise inpainting} fixes selected body parts while allowing the model to infer the remaining joints, such as preserving upper-body motion while completing the lower-body dance, or vice versa. Third, \textbf{trajectory-guided inpainting} constrains sparse trajectory-related channels such as root translation or turning direction, and lets the model generate the full-body pose sequence that follows the prescribed path. These examples show that \textbf{MACE-Dance} is not limited to one-shot motion generation, but can also function as a controllable motion editing tool for choreography and animation workflows.

Another important advantage of the proposed \textbf{Motion Expert} is that its output is explicit 3D motion, which can be directly transferred to standard character rigs through conventional motion retargeting pipelines, as also shown in Fig.~\ref{fig:supp_motion_editing}. This substantially broadens the applicability of \textbf{MACE-Dance} beyond music-driven video synthesis. In addition to being rendered by our \textbf{Appearance Expert}, the generated dance motion can be reused as a structured motion asset for CG animation, VR avatars, interactive character control, and other human-computer interaction scenarios that require editable and transferable body motion. More broadly, because the output remains in a structured 3D form, the framework is also potentially extensible to embodied platforms such as humanoid agents or dancing robots after appropriate skeleton mapping and control-level adaptation. In this sense, the \textbf{Motion Expert} is not only a component for improving video generation quality, but also a general-purpose music-to-motion generator with strong downstream utility in animation, XR, and embodied AI applications.

% \begin{figure}[t]
%     \centering
%     \includegraphics[width=0.95\linewidth]{figs/supp_motion_editing.pdf}
%     \caption{\textbf{Motion editing and retargeting results of the Motion Expert.} Using masked denoising, the proposed model supports temporal inpainting, joint-wise inpainting, and trajectory-guided inpainting by preserving user-specified motion constraints and completing the unknown regions with plausible, music-aligned 3D dance motion. The generated motion can also be retargeted to CG characters, demonstrating its utility beyond video synthesis in applications such as animation, VR avatars, and embodied character control.}
%     \label{fig:supp_motion_editing}
% \end{figure}

\section{Further Discussion about MACE-Dance}
% \subsection{3D-to-2D Motion Projector}
% The 3D-to-2D Motion Projector module is designed to convert 3D SMPL parameters, into a 2D keypoint sequence. This projection is crucial for bridging the domain gap between the 3D parametric representation and the 2D structural information required by our downstream network components. The process unfolds in two main stages. 
% First, for each frame of a 3D motion sequence, we convert its SMPL parameters into a 3D mesh. We then employ an offscreen renderer (\texttt{pyrender}) with a fixed-perspective camera to generate a canonical frontal view image of the mesh. 
% Second, the resulting sequence of rendered images is processed by a powerful, pre-trained 2D pose estimator, \texttt{ViTPose}~\cite{xu2022vitpose}, which is compatible with the pose input specification of WAN-Animate. This widely adopted model analyzes each image to extract the 2D coordinates of human body joints. 

\subsection{Temperature Parameter $\beta$ in GFT}
As mentioned in main paper Sec. 3.2.1, we adopt Guidance-Free Training (GFT~\cite{chen2025visual}). GFT reformulates conditional training to directly learn a $\beta$-indexed sampling model via linear interpolation with the unconditional output. This allows a single model to capture an entire family of diversity-fidelity trade-offs robustly, eliminating the need for post-hoc guidance. $\beta$ serves as a temperature parameter that is also provided to the model $\theta$ as an additional conditioning input. During inference, values of $\beta$ near 0 favor high fidelity, while values near 1 favor high diversity. Thus, $\beta$ can also be regarded as a control signal, and we set its value to 0.75.
To empirically validate the effect of the parameter $\beta$ and justify our choice, we conducted an ablation study as presented in Tab.~\ref{tab: beta}. The results confirm the expected trade-off between diversity and fidelity. Specifically, $\beta=1.00$ yields the highest diversity scores ($DIV_k=13.29, DIV_g=9.68$) but suffers from the poorest fidelity. Conversely, $\beta=0.50$ achieves the best fidelity ($FID_k=15.11, FID_g=24.15$) at the expense of diversity, which drops below the ground truth. A value of $\beta=0.00$ leads to numerical instability, confirming it is unsuitable for inference. We ultimately select $\beta=0.75$ as it offers the most compelling balance. It dramatically improves fidelity over $\beta=1.00$ (e.g., $FID_k$ drops from 29.35 to 17.83) while retaining strong diversity ($DIV_k=10.30, DIV_g=8.09$) that surpasses both the high-fidelity setting ($\beta=0.50$) and the ground truth. This makes it the optimal choice for producing results that are both high-quality and varied.

\begin{table}[!t]
\centering
\renewcommand{\arraystretch}{1.2}
\setlength{\tabcolsep}{5pt}
\caption{Effect of the $\beta$ in Guidance-Free Training (GFT).}
% \vspace{-0.1in}
\label{tab: beta}
\scriptsize % 在这里设置字号
\resizebox{0.9\linewidth}{!}{ % 宽度可以根据需要调整
\begin{tabular}{l|cccccc} % 只在第一列后保留竖线
\toprule
 $\beta$ & FID$_{k}$$\downarrow$ & FID$_{g}$$\downarrow$ & FSR$\downarrow$
 & DIV$_{k}$$\uparrow$ & DIV$_{g}$$\uparrow$
 & BAS$\uparrow$ \\ 

\midrule
\textit{GT}          
& --     & --     & 0.216
& 9.94 & 7.54 & 0.201 \\

1.00 & 29.35 & 31.91 & 0.270
& \textbf{13.29} & \textbf{9.68} & 0.220 \\

0.75
& 17.83 & 25.09 & \textbf{0.210}
& 10.30 & 8.09 & 0.229 \\

0.50
& \textbf{15.11} & \textbf{24.15} & \textbf{0.210}
& 8.64 & 6.79 & \textbf{0.233}\\

0.00
& NaN & NaN & NaN
& NaN & NaN & NaN \\

\bottomrule
\end{tabular}
}
% \vspace{-0.2in}
\end{table}

\subsection{Task Decoupling Analysis}
MACE-Dance is a music-driven dance video generation framework with a cascaded Mixture-of-Experts (MoE) architecture, which decouples the task into music-to-3D motion generation (Motion Expert) and pose-driven image animation (Appearance Expert). This design is motivated by the principles of reducing learning complexity and improving data utilization, as detailed below:

\textbf{(1) Complexity reduction via task factorization.}
By separating the original cross-modal mapping from music directly to pixels into two more constrained subproblems, each expert can focus on a well-defined objective. The Motion Expert specializes in modeling the temporal relationship between music and human kinematics, without interference from visual factors such as texture or lighting. Conversely, the Appearance Expert addresses a conditional image synthesis task given explicit pose inputs, without requiring an understanding of musical semantics. This specialization enables each expert to learn a more robust and domain-appropriate representation.

\textbf{(2) Suppression of spurious cross-modal correlations.}
End-to-end models are prone to learning incidental correlations between musical features and visual artifacts present in the training data (e.g., background or clothing cues). Introducing an explicit 3D motion representation acts as a structured information bottleneck, compelling the model to focus on the intrinsic relationship between music and movement while filtering out irrelevant visual factors. We empirically observe this phenomenon when adapting several representative end-to-end human motion generation models, including \texttt{Hallo2}, \texttt{EchoMimic-V3}, and \texttt{WAN-S2V}. Despite architectural modifications or fine-tuning, these models exhibit clear spurious correlations. This limitation is reflected in the consistent performance gap between these baselines and our method, as reported in Tab.~1, Fig.~3, Fig.~8 of the main paper, and Fig.~4 in the appendix.

\textbf{(3) Interpretability and explicit control through structured representations.}
The intermediate 3D motion representation provides a transparent and editable interface that can be inspected, modified, or replaced prior to final rendering. Such interpretability and controllability are fundamentally unavailable in monolithic end-to-end models. Overall, the cascaded MoE design facilitates model specialization, improves data efficiency, and enables user-level control, leading to more robust and reliable dance video generation.

\subsection{Long-Sequence Generation}
In the domain of dance video generation, long-sequence generation is not merely an enhancement but a fundamental requirement for practical applications. Its importance is multifold: first, a complete dance performance is an expressive narrative with an emotional arc, intrinsically tied to the full duration of a musical piece (typically 30s-4min). Short clips fail to capture the choreographic structure, narrative progression, and full artistic integrity. Second, to achieve precise music synchronization, the model must process motion sequences matching the entire length of the musical score, ensuring long-term alignment of movements with the beat, melody, and mood. However, prevailing methods in general human video generation are often constrained by the limited temporal window of their underlying base models~\cite{peng2024synctalk,peng2025omnisync,peng2025synctalk++} (e.g., under 5 seconds). Naively extending these models to long-sequence tasks inevitably confronts the critical challenge of error accumulation. This error manifests as motion drift, identity degradation, and temporal incoherence. To overcome this core problem, our framework employs a synergistic two-stage strategy, achieving high-quality long-sequence dance video generation, as shown in Fig. ~\ref{fig: supp_longdance}:

\begin{figure}[!t]
  \centering

    \includegraphics[width=0.95\linewidth]{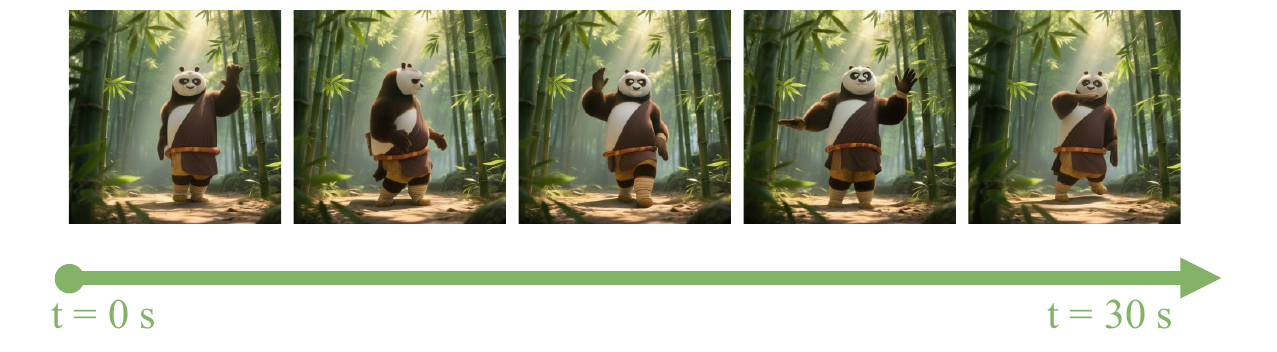}
    % \vspace{-0.2in}
    \caption{\textbf{MACE-Dance} produces long-sequence dance videos with artistic expressiveness and physical plausibility.}
    % \vspace{-0.1in}
    \label{fig: supp_longdance}
  % \end{subfigure}
\end{figure}

\textbf{(1) Motion Expert with length extrapolation capability.}
The Motion Expert employs a BiMamba–Transformer hybrid architecture that combines global structural modeling with local temporal continuity. Transformer blocks capture global choreographic structure and long-range dependencies via self-attention, while BiMamba layers model local motion dynamics with linear complexity.
Although trained on short motion clips (e.g., 8 seconds), the model can generate sequences of arbitrary length at inference time. This is enabled by the state-space recurrence of Mamba, which serves as a temporal memory that continuously propagates local dynamics beyond the training horizon, while the Transformer provides high-level structural guidance within its receptive field.

\textbf{(2) Pose-anchored relay generation in the Appearance Expert.}
Given the coherent long motion sequence, the Appearance Expert renders the final video using a pose-driven image animation paradigm rather than generic video prediction. Each generation chunk is constrained by three complementary anchors: (i) the globally consistent 2D pose sequence from the Motion Expert, which provides an absolute geometric reference; (ii) the last frame of the previous chunk, ensuring appearance continuity (e.g., lighting and clothing); and (iii) a constant reference image, enforcing identity consistency. Together, these constraints effectively prevent error accumulation and maintain long-term visual coherence, in contrast to unconstrained autoregressive video generation.

\section{Ethical Considerations}
Although \textbf{MACE-Dance} is designed for music-driven dance video generation in creative and entertainment contexts, it may also introduce ethical risks. 
In particular, as with other human video generation systems, the model could be misused to synthesize realistic videos of individuals without their consent, potentially enabling misleading or deceptive media. 
This concern is especially relevant because the \textbf{Appearance Expert} preserves identity-related cues from a reference image while generating temporally coherent videos.

Moreover, the training data may exhibit biases in dance style, body shape, clothing, scene composition, and cultural representation, which can lead to uneven generation quality across different subjects or styles. 
Accordingly, the outputs of the model should not be interpreted as neutral or universally representative.

We stress that \textbf{MACE-Dance} is intended for research on controllable dance video synthesis, not for identity manipulation or harmful content creation. 
Any practical deployment should respect consent, portrait rights, and copyright constraints, and future releases should consider safeguards such as usage restrictions, provenance disclosure, or watermarking mechanisms.

\section{Limitations and Future Work}
\subsection{Customized Dance Generation.}
Although our framework \textbf{MACE-Dance} achieves strong performance in music-driven dance video generation. Music serves as a fixed-form carrier and cannot fully capture diverse user intentions. To address the limitations, we envision extending control modalities to incorporate free-form textual descriptions.
Text offers the lowest-cost input modality while allowing users to express choreographic requirements in a more flexible and semantically rich manner, thereby facilitating personalized and expressive dance generation. 
Specifically, text provides a rich, hierarchical control mechanism, enabling users to articulate dance from multiple levels of abstraction. It can define high-level artistic concepts like mood and style (e.g., 'an energetic hip-hop dance'), while also specifying low-level kinematic details such as a sequence of actions or the movement of a particular limb (e.g., 'spin and then raise both arms'). 
This direction not only enhances user interactivity and creativity but also unlocks new opportunities for content-driven applications in human–computer interaction. 
While recent studies have explored text-controlled human video generation~\cite{peng2025actavatar}, current approaches are hindered by the limited scale of available dance video and the difficulty in acquiring textual descriptions that not only align with natural user expression patterns but also precisely reflect the essential characteristics of dance movements. 
Thus, leveraging text as a control modality is a pivotal next step, promising to unlock truly personalized and creative dance generation.

\subsection{Dance Generation with Efficiency.}
Real-time interaction represents a critical and compelling direction for dance video generation. While our \textbf{Motion Expert} has achieved state-of-the-art (SOTA) generation efficiency in the 3D motion synthesis stage, a significant performance bottleneck remains in the \textbf{Appearance Expert}. 
Specifically, although our fine-tuned \textbf{Appearance Expert}, based on the 14B-parameter \texttt{Wand-Animate} model, delivers SOTA quality in pose-driven image animation, its substantial computational demands preclude its use in real-time applications.
To bridge this efficiency gap, several promising research avenues can be explored. These include \textbf{knowledge distillation}, where a compact student model is trained to mimic the large teacher model; \textbf{model compression} techniques like quantization and pruning; and, more fundamentally, designing a novel, lightweight \textbf{Appearance Expert} architecture optimized for speed. Ultimately, achieving a harmonious balance between generation quality and computational efficiency is the key to unlocking the full potential of interactive dance video synthesis.

\end{document}